\documentclass{scrarticle}
\usepackage[utf8]{inputenc}
\usepackage[T1]{fontenc}
\usepackage[a4paper, left=2cm, top=2cm, right=2.5cm, bottom=2cm]{geometry}
\usepackage{xcolor}
\usepackage{tikz}
\usepackage[sort]{natbib}
\usepackage{amssymb}
\usepackage{amsmath}
\usepackage{bm}
\usepackage{tabularx}
\usepackage{subcaption}
\usepackage{hyperref}
\usepackage[colorinlistoftodos, textsize=tiny]{todonotes}
\usepackage{tablefootnote}
\usepackage{tipa}
\usepackage{siunitx}
\usepackage{booktabs}
\usepackage{authblk}
\usepackage{datetime}
\newdateformat{monthyeardate}{%
  \monthname[\THEMONTH], \THEYEAR}
\usepackage{CJKutf8}
\newcommand{\mandarin}[1]{\begin{CJK*}{UTF8}{bsmi}#1\end{CJK*}}

\title{Word-specific tonal realizations in Mandarin}
\author[1]{Yu-Ying Chuang}
\author[2]{Melanie J. Bell}
\author[3]{Yu-Hsiang Tseng}
\author[3]{R. Harald Baayen}
\affil[1]{National Taiwan Normal University, Taiwan}
\affil[2]{Anglia Ruskin University, UK}
\affil[3]{Eberhard Karl University of Tübingen, Germany}

\date{\monthyeardate\today}

\begin{document}

\maketitle

\begin{center}
  \textbf{Abstract}
\end{center}

\noindent
The pitch contours of Mandarin two-character words are generally understood as being shaped by the underlying tones of the constituent single-character words, in interaction with articulatory constraints imposed by factors such as speech rate, co-articulation with adjacent tones, segmental make-up, and predictability. This study shows that tonal realization is also partially determined by words' meanings. We first show, on the basis of a corpus of Taiwan Mandarin spontaneous conversations, using a generalized additive regression model, and focusing on the rise-fall tone pattern, that after controlling for effects of speaker and context, word type is a stronger predictor of tonal realization than all the previously established word-form related predictors combined. Importantly, the addition of information about meaning in context improves prediction accuracy even further. We then proceed to show, using computational modeling with context-specific word embeddings, that token-specific pitch contours predict word type with 50\% accuracy on held-out data, and that context-sensitive, token-specific embeddings can predict the shape of pitch contours with 40\% accuracy. These accuracies, which are an order of magnitude above chance level, suggest that the relation between words' pitch contours and their meanings are sufficiently strong to be potentially functional for language users. The theoretical implications of these empirical findings are discussed. \\ \ \\

\noindent
\textbf{keywords}: Tone, Mandarin, embeddings, GAMs, form-meaning isomorphy, rise-fall tone pattern, two-syllable words  \\
\ \\
\ \\
\ \\
\ \\
\noindent
\textbf{Acknowledgements}
This research was funded by the European Research Council, grant SUBLIMINAL (\#101054902) awarded to Harald Baayen.  The authors are indebted to Dr. Matteo Fasiolo, University of Bristol, for his statistical advice on the application of GAMs to pitch contours.
\ \\
\ \\
\ \\
\noindent
\textbf{Author note}
Correspondence concerning this article should be addressed to Dr.  Yu-Ying Chuang, National Taiwan Normal University, e-mail: yuying.chuang@ntnu.edu.tw.

\newpage

\section{Introduction}
\label{sec:introduction}

The tone system of Mandarin Chinese is commonly described as having four lexical tones, namely high level (T1), rising (T2), dipping/low (T3), and falling (T4), plus one neutral or floating tone whose shape depends on the preceding tone \citep{chao1968grammar}. While these tonal representations are well established in the relevant literature, and are taught both in Chinese schools and in second language learning classrooms, it is also known that their realizations, i.e. the actual pitch contours of spoken words, can differ significantly from the canonical descriptions.

In fact, the mismatch between words' actual realizations and their underlying forms is ubiquitous. Substantial variability can be observed when one investigates how a word is pronounced in spontaneous speech. \citet{Johnson:2004}, for example, reports that in English conversational speech, the word `until' is pronounced as [\textipa{2ntIl}], [\textipa{2nt@l}], [\textipa{EntIl}], [\textipa{Ent@l}], [\textipa{IntIw}], [\textipa{\s{n}tIl}], [\textipa{@nt2}], [\textipa{\s{n}t\s{l}}], [\textipa{t\s{l}}], and [\textipa{t@}]. That is, a word represented by a single orthographic form can undergo massive syllable deletion and segment deviation, leading to multiple pronunciation variants. The same holds for Mandarin. \citet{chung2006contraction} observed that the connective \mandarin{雖然} [\textipa{s\super weI\:zan}] `although' is often pronounced as [\textipa{s\super we\*ra}] or [\textipa{s\super weI\s{m}}], and \mandarin{可是} [\textipa{k\super h@\:s1}] `but' as [\textipa{k\super h@\:R}] or simply [\textipa{k\super h@:}]. About one-third of the syllables produced in spontaneous Mandarin are contracted syllables, i.e., they are reduced pronunciations of multi-syllabic words \citep{Tseng2005SyllableCI}. It can thus be seen that 
the phonetic realizations of a word vary, and the canonical description of a word's phonological form is rarely encountered in conversational speech.

Not only do tokens of the same word vary substantially in the realization of their segments but, in tonal languages, they also vary substantially in the realization of their lexical tone. \citet{xu2001sources} distinguishes two sources of tonal variation: voluntary and involuntary. Voluntary variation arises from the speaker's communicative intentions, such as eliciting a question or creating emphasis, where changes in rhythm, accent placement, and intonation can all give rise to substantial modification of tonal realization; these voluntary constraints are intimately related to the syntactic and pragmatic functions of an utterance
\citep{gaarding1987speech, shen1989interplay, shen1990prosody, xu1999effects, liu2005parallel}. In addition, the realization of tone is modulated by sociolinguistic and paralinguistic factors such as gender, dialect, and emotion \citep{fon1999does,zhang2006acoustic}. 

Involuntary variation in tone production arises from articulatory constraints posited to be beyond the speaker's control. Some of these constraints are related to the processes of connected speech. At the utterance level, where a sequence of tones is produced, the realization of a given tone is greatly influenced by its preceding and following tones, leading to tonal co-articulation \citep{shih1988tone, shen1990tonal, xu1997contextual}. Co-articulated tones usually deviate from their canonical tonal shapes; in extreme cases the original shapes are no longer preserved \citep{shen1989interplay, shih2000chinese} and may be unrecognizable to native speakers \citep{xu1994production}. Physically speaking, it takes a certain amount of time to raise or lower pitch \citep{xu2002maximum}, and therefore tonal realizations are highly dependent on whether speakers have sufficient time to realize a given tonal contour. Under the time pressure of fast speech, tonal targets can usually not be fully realized \citep{tang2020acoustic}; this leads to significant deviation from the canonical patterns and often results in tonal reduction \citep{cheng2015mechanism}. The accepted descriptions of the tones as `level', `rising', `dipping/low', and `falling' are therefore generalizations across considerable variability in the fine detail of their phonetic realizations.

In addition to the effects of connected speech, articulatory constraints on tone production arise from the segmental makeup of syllables. At the syllable level, vowels, onsets, and syllable structure are all known to contribute to tonal variation \citep{howie1974domain, ho1976acoustic, whalen1995universality, xu2003effects, fon2007positional}. Over and above contextual effects, it is therefore clear that different words with the same canonical tone will differ in the details of their tonal contours, since they differ in their segmental makeup. In previous studies on Mandarin lexical tones using laboratory speech, by-word variation is rarely taken into account, since very often the same words are used across different experimental conditions for maximum control \citep[e.g.,][]{chen2010post, li2016acoustic}. Previous studies using corpus data have accommodated by-word variation as a random effect in the relevant statistical models. For example, \citet{wu2020mandarin} worked with random intercepts for word-form. However, this treats the word as a source of noise, where different words exhibit idiosyncrasies that are irrelevant to the predictors of interest. In contrast, other studies have specifically investigated how word-level properties affect tone production. For example, \citet{zhao2009effect} showed that usage frequency affects F0 realization in Cantonese words with mid-level or mid-rising tones.\footnote{\citet{zhao2009effect} describe Cantonese as having six lexical tones.} For Mandarin, \citet{tang2021prosody} showed that F0 is affected not only by word frequency but also by informativity, defined as a word-level variable. However, to the best of our knowledge, no previous study has specifically addressed the relationship between tonal realization and word meaning. This paper fills that gap.

In the extensive literature on Mandarin tones, their semantic function is straightforwardly simple: different tones distinguish between alternative meanings. For instance, \mandarin{就} \textit{jiu4} `then' and \mandarin{九} \textit{jiu3} `nine' are differentiated by having a falling and a dipping tone, respectively. However, the very same combinations of segments and tone often realize many other different meanings, as exemplified by \mandarin{九} \textit{jiu3} `nine' and \mandarin{酒} \textit{jiu3} `alcoholic beverage'.\footnote{Throughout this paper, pinyin transcriptions are followed by numeric notations representing the four tones.} The combination of strong phonotactic constraints on syllable structure and a limited number of lexical tones has given rise to widespread homophony and polysemy, often in combination with homography. For instance, \mandarin{就} \textit{jiu4} `then', has a wide range of translation equivalents in English, including \textit{then, at once, only, already, to approach, to accomplish, to suffer}, and \textit{to take advantage of} (\url{https://www.pleco.com/}, s.v.).

The examples in the preceding paragraph are all monosyllabic words; however, in the Chinese Lexical Database \citep{sun2018chinese}, only 8\% of the 48,000 words are monosyllabic. The majority (70.4\%) of Mandarin words are disyllabic, written with two characters.\footnote{In this study we use a corpus of Taiwan Mandarin spontaneous speech \citep[cf.][]{fon2004} and take the word labels supplied by the corpus as given. The labeled words include, for example, nouns such as \mandarin{學校} {\em xue2xiao4} `school', verb forms such as \mandarin{學到} {\em xue2dao4} `learn+resultative', and negated verbs such as \mandarin{不是}  {\em bu2shi4} `not+be'. Since all these forms have two syllables, we refer to them collectively as disyllabic words.} The tonal targets of disyllabic words are taken to be subject to the same voluntary and involuntary constraints that govern the realization of tone in monosyllabic words. As a consequence, all disyllabic words sharing, for example, an initial falling tone and a subsequent rising tone are assumed to have the same underlying pitch contour; any differences in how the tones are realized  are assumed to be attributable to the involuntary and voluntary processes described above. However, the present study will show that, alongside the known articulatory constraints,  there is a previously undocumented close association between the meanings of Mandarin disyllabic words and the realization of their tonal contours.

The basis for our study was laid by a growing body of research on English showing that fine-grained phonetic variation can be systematically associated with differences in meaning. For example, at the word level, \citet{Gahl:2008} showed that homophones such as {\em time} and {\em thyme} are realized with different acoustic durations. In the same vein, \citet{lohmann2018cut} found that the durations of words such as {\em cut} depend on whether they are used as nouns or verbs. These differences in word duration were initially explained as a consequence of the different relative frequencies of the homophones in these studies. However, \citet{gahl2022time} showed that the meanings of English homophones are a strong co-determinant of their spoken word durations even after frequency differences and other co-determinants such as speech rate are taken into account. A relationship between meaning and duration has also been found for the English suffix /s/: different grammatical functions of this suffix (e.g., plural and third person singular) tend to be realized with different durations \citep{plag2017homophony}. Furthermore, the relationship between meaning and phonetic realization may extend beyond durational differences. \citet{drager2011sociophonetic} reported that the phonetic realization of the word {\em like} varies according to its discourse or grammatical meanings, not only in the duration of the consonants but also in the degree of diphthongisation of the vowel. 

The results described in the previous paragraph are compatible with a theory of the mental lexicon that postulates a direct connection between the context-specific meaning of a word token and the details of its phonetic realization. Such a theory has been computationally implemented as the Discriminative Lexicon Model, henceforth \textbf{DLM} \citep{baayen2019discriminative,Chuang:Baayen:2021,Heitmeier:Chuang:Baayen:2024}. In this model, lexis and morphology are acquired through a process of error-driven learning that allows for fine-grained alignments between low-level properties of form and low-level properties of meaning, both operationalized as high-dimensional numeric vectors. The model captures relationships between these vectors in two networks: a comprehension network that maps word form onto word meaning, and a production network that maps word meaning onto word form. 
In other words, given a word form vector, the comprehension network is able to compute a predicted meaning vector, analogous to the comprehension process in which we recognize and make sense of the meaning of a word according to the visual or auditory input. In the same vein, the production network computes a predicted form vector given a semantic vector, analogous to the production process where we produce a form to express an intended word meaning. It has been shown that, across a range of languages, these networks successfully capture an alignment between meaning and fine-grained variation in form, e.g., word duration in Mandarin \citep{Chuang:Kang:Luo:Baayen:2021}, different durations of homophones in English \citep{gahl2022time}, and the degree of tongue lowering in articulation of the [a] vowel in German \citep{saito2022articulatory}.

Despite being able to predict both word forms and word meanings, the DLM model does not store whole word representations of either kind; rather, the model's memory consists of the connection weights in the networks, which are continuously re-calibrated with each learning event. Likewise, in the corresponding theory of the mental lexicon, word forms and meanings do not have representations in memory. The forms are ephemeral auditory or visual experiences, which dynamically generate corresponding, equally ephemeral, meaning representations. Conversely, a meaning conceptualized by a speaker at a given point in time is dynamically transformed into ephemeral representations driving articulation. In other words, the DLM posits a lexicon in which lexical items are neither static nor discrete. Rather, the lexicon is taken to consist of a series of dynamic, modality specific neural networks \citep{baayen2019discriminative} which are constantly fine-tuned, by adjusting connection weights, in order to optimize word comprehension and production. For further technical details of the implementation of the DLM mappings, please refer to the supplementary material.

At this point, the question arises of how to understand the linguistic term `word'. In this study, we define \textbf{word token} as a pairing of a specific form with a specific meaning. We define  \textbf{word type} (or simply, \textbf{word}) as a set of word tokens that have the property of having both similar forms and similar meanings. For instance, the set of phonetic realizations of \mandarin{酒} \textit{jiu3} and their corresponding context-specific meanings (`wine, liquor, spirits, alcoholic beverage', \url{https://www.pleco.com/}, s.v.) jointly constitute the tokens of the word type \mandarin{酒}.\footnote{Note that, in this case, the Chinese orthographic character can be used to represent the word, since \mandarin{酒} is not homographic: all its meanings cluster around the concept of `alcoholic beverage'.}  This working definition of `word' seeks to do justice to the fact that no two tokens of the same word, as produced by humans, are ever completely identical in form. It also seeks to do justice to the insights from distributional semantics that what a word means varies with its context \citep{Firth:1968,Harris1954,Landauer:Dumais:1997,Elman:2009dinosaur}.
Thus, in the framework of the DLM, words are sets of input-output pairs on which the production and comprehension networks are trained, that leave `traces' in the connection weights of these networks, but that are themselves not stored as independent entities. 

As may be apparent from the preceding paragraph, our theoretical position apropos meaning is one of contextualism. We assume that utterances rather than sentences are the domain of propositional content, and that the meaning of an utterance depends on the context in which it is produced. Our notion of context includes not only the content, genre and style of the surrounding text but also extralinguistic factors such as who is speaking and who they are addressing. Hence, our view of meaning does not draw a neat boundary between semantics and pragmatics. 

Both the theoretical assumptions of the DLM and the empirical studies of English homophonous words and affixes by \citet{gahl2022time} and \citet{plag2017homophony} respectively, suggest the possibility that Mandarin homophones also differ systematically in phonetic detail, i.e., that their segments and/or tones are realized slightly differently according to the intended meaning. This could apply to homographic pairs such as \mandarin{大家} \textit{da4jia1} `everyone' and `art master', as well as to non-homographic pairs such as \mandarin{樹木} \textit{shu4mu4} `tree' and \mandarin{數目} \textit{shu4mu4} `number'. In other words, it is possible that the realizations of canonical tones are determined not only by the involuntary and voluntary constraints previously described, but also by the context-specific meanings of the word tokens on which they are realized. If this is correct, then conversely it is not only the four canonical pitch contours that help to distinguish between alternative meanings, but also the finer details of their phonetic realization.  This brings us to the central hypothesis of our study:

\begin{itemize}
\item[]\textit{The unique pitch contour of each spoken Mandarin word token is determined in part by the specific meaning of that token. These meaning-specific contours are in principle learnable and hence potentially used by speakers.}
\end{itemize}

\noindent
In what follows, we provide evidence for four more specific predictions that we derive from this hypothesis:
\begin{enumerate}
   \item Word type will be a stronger predictor of tonal realization than all the previously established word-form related predictors combined. This prediction follows from the hypothesis because word type includes information about meaning in addition to information about form.   
  \item Information about a word's meaning in context, i.e. its sense, will improve prediction of its tonal realization, compared with prediction based on word type alone. This prediction follows from the hypothesis because a word's senses include all the information encoded in the word, plus additional more fine-grained semantic information. It therefore allows us to tease apart effects of meaning from possible effects of form.
  \item Given a pitch contour, the meaning of its carrier token can be predicted above chance level by a simple computational model with previous experience of that word type. This prediction follows from the hypothesis that meaning-specific contours are in principle learnable.
   \item Assuming it has previous experience of the relevant word type, a simple computational model can produce an appropriate pitch contour for a given meaning. This also follows from the hypothesis that meaning-specific contours are in principle learnable.
\end{enumerate}

\noindent
In this paper we explore these predictions for disyllabic words with the canonical tone specification of a rising tone (T2) followed by a falling tone (T4), henceforth \textbf{RF}. The disyllabic word is a natural choice for our study since Mandarin vocabulary is composed mostly of disyllabic words \citep{Huang2010, wu2023mandarin}. We decided to focus on the RF pattern, because it is the heterogeneous tonal combination with the highest number of word types and tokens in the speech corpus we used. We wanted to investigate a heterogeneous tonal combination rather than a homogeneous one to ensure that the results obtained are not specific to a given tone. 

The remainder of the paper proceeds as follows. Section \ref{sec:pitch} addresses the first two predictions listed above. It describes how we used generalized additive modeling to analyze the pitch contours of RF words extracted from the above-mentioned corpus of spontaneous speech. We discuss our modeling strategy, and present the results of an analysis based on word type and one enhanced with word sense. Section \ref{sec:model} addresses the third and fourth predictions. It describes how we used computational modeling with the DLM  to demonstrate that meaning-specific pitch contours have the potential to facilitate comprehension and to be produced in response to intended meaning. Section \ref{sec:discussion} is a discussion of the implications of our results.

\section{Establishing word and meaning-specific pitch contours} \label{sec:pitch}

\noindent
This section describes how we addressed the first two predictions outlined in Section \ref{sec:introduction}. We modeled the pitch contours of spoken tokens of Mandarin disyllabic words with the RF tonal pattern, using generalized additive modeling. To explore Prediction 1, we evaluated the effectiveness of word type as a predictor of tonal realization, compared with the segment-related articulatory constraints previously described in the literature. To explore Prediction 2, we evaluated whether adding information about a word's meaning in context would improve prediction of tonal realization, compared with prediction based on word type alone. 
Sections \ref{sec:methodology} to \ref{sec:strat} describe aspects of the methodology, Sections \ref{sec:baseline} to \ref{sec:sense_id} report the results of the generalized additive models, and Section \ref{sec:discuss_gam} summarizes Section \ref{sec:pitch} overall.

\subsection{Generalized additive modeling} \label{sec:methodology}

Classical analyses of pitch typically take measurements at various contour landmarks, such as maximum and minimum F0 values. However, since pitch actually varies continuously with time, such analyses miss much of the detail of the F0 contour. To better capture the complete shapes of tonal variations, we modeled the pitch contours of the tokens in our data using generalized additive models, henceforth \textbf{GAM}s \citep{Wood:2017}. GAMs relax the regression assumption that the relation between a predictor and response should be linear; instead, the model incorporates individual, potentially nonlinear relationships between each predictor variable and the response variable. For main effects, this relationship is estimated using functions known as smoothing splines (henceforth \textbf{smooths}), which can fit either a line or a (possibly wiggly) curve to the data, as required. Nonlinear interactions can be included using functions called \textbf{tensor product smooths}, which fit a wiggly (hyper)surface for the joint effect of two or more predictors. In addition, it is possible to include nonlinear random effects, for instance by using functions called \textbf{factor smooths} \citep{baayen2022note}, which fit a wiggly curve for each level of the random factor, e.g. for each individual speaker in the case of a by-speaker factor smooth. Because GAMs model complex non-linear relationships, they make it possible to model F0 as a nonlinear function of time across an utterance, while also including other predictors known to affect pitch, such as speech rate and speaker gender. They can thus capture fine-grained modulations of pitch as time unfolds.

\begin{figure}
  \centering
  \includegraphics[width=0.4\textwidth]{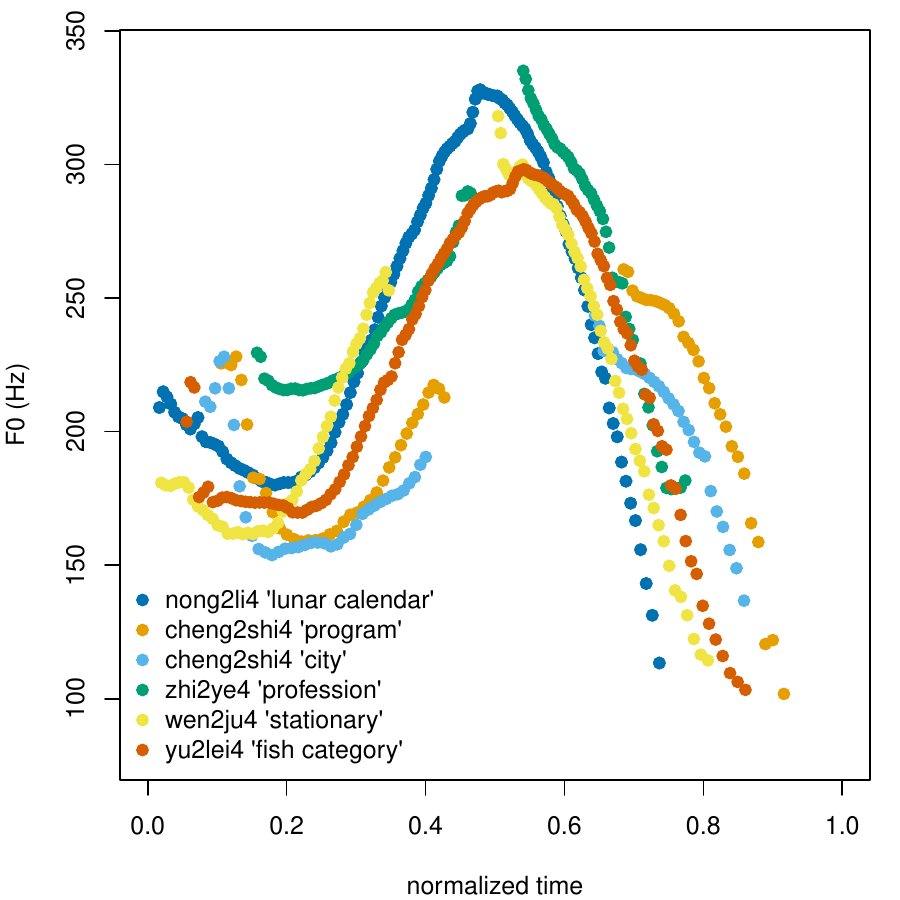}
  \hspace{-0.08cm}
  \includegraphics[width=0.4\textwidth]{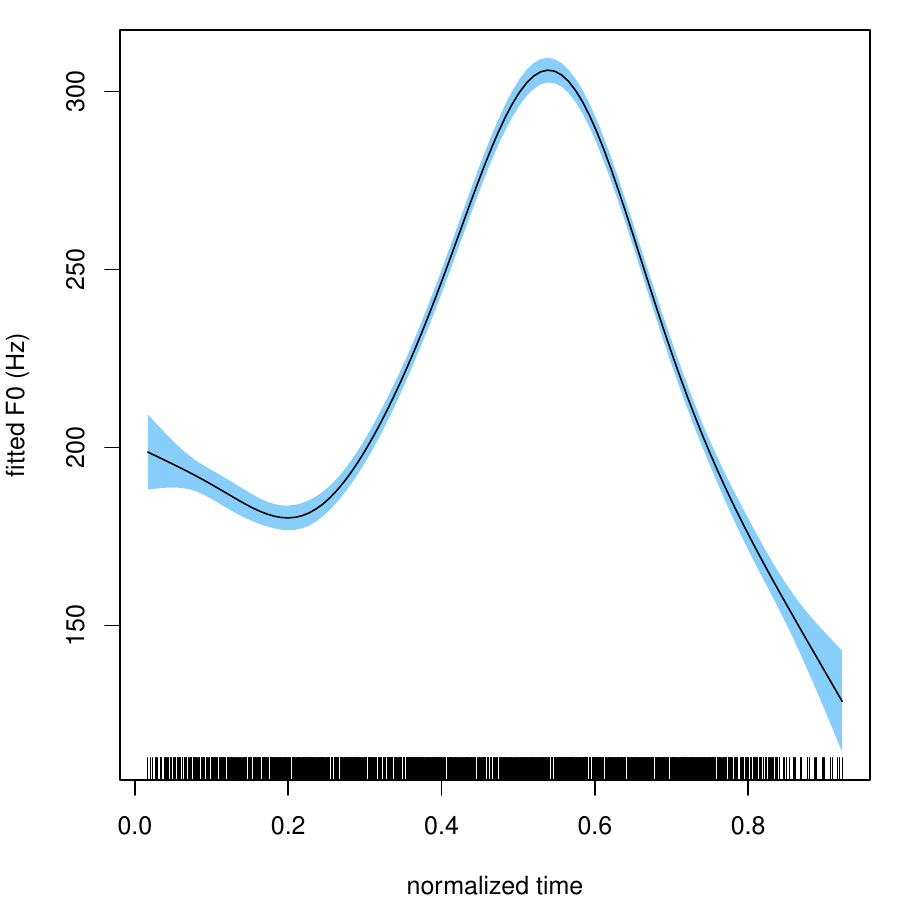}
  \caption{Toy dataset. The left-hand panel shows the F0 contours of single tokens of six Taiwan Mandarin words with the RF tonal pattern, produced in isolation by the same speaker. The right-hand panel shows the RF contour predicted by the a simple GAM, using a thin plate regression spline smooth for normalized time as predictor. }
  \label{fig:toy_contour}
\end{figure}

To illustrate our GAM-based modeling strategy, we created a toy dataset consisting of six disyllabic Mandarin word types with the RF tonal pattern. Their audio files were downloaded from {\em Meng Dian}, a publicly available Taiwanese online dictionary. In the audio files, a single female speaker pronounces each word twice, so that we had a total of 12 tokens, all from the same speaker. We used the \textsc{To Pitch (cc)} command in Praat \citep{praat} to estimate F0 values for each token. Because the speaker is female, we set the pitch floor at 50 Hz and the pitch ceiling at 400 Hz. To optimize the F0 estimation, we set the time step at 0.001 seconds and used the most accurate method available, namely a Gaussian window; other parameters were left at the default values. This gave us F0 values at one millisecond intervals for the voiced sections of the tokens. Next, we used Praat's \textsc{To PointProcess} command to obtain the time points of the glottal pulses in the voiced sections of the tokens. We then extracted the F0 values corresponding to the time points of the glottal pulses. 
Since the time points of glottal pulses do not necessarily correspond exactly to the one-millisecond measurement intervals, we used linear interpolation between adjacent measurements to estimate F0 for the glottal pulses. 
At this stage no F0 values were interpolated for the voiceless sections of the tokens. Because the tokens varied in duration, we transformed the time points of the measurements onto a normalized time scale of 0-1.\footnote{
Time normalization enables us to focus on the shape of pitch contours.  In order to take into account possible effects of word duration on the realization of tone, in the more comprehensive analyses reported below, token duration is included as a covariate, in interaction with normalized time. Note that by reversing the time normalization, the predicted curves can straightforwardly be back-transformed to the original time scale.}
The F0 values of one of the two renditions of each of the six words are plotted on this normalized time scale in the left-hand panel of Figure~\ref{fig:toy_contour}. Although there are no data points for the voiceless segments, it can be seen that the RF tonal sequence in Mandarin is realized with a small initial fall, followed by a rise, and finally a much larger fall. The dipping realization of T2 is consistent with previous findings for laboratory speech that the rising portion of Mandarin T2 is usually preceded by a slight fall \citep{ho1976acoustic, tseng1981acoustic, shih1988tone, shen1991perceptual, xu1997contextual, moore1997speaker}.\footnote{The initial fall could also reflect dialectal variation specific to Taiwan Mandarin, as in this language, T2 is predominantly realized with a concave contour. This concave contour may have become a standardized realization that no longer reflects articulatory constraints \citep[see e.g.,][]{fon2007positional}.}

Using the \textbf{mgcv} package \citep{Wood:2017} for R \citep{Rcore2022}, we fitted a GAM to the toy dataset, with F0 as the dependent variable and normalized time as the only predictor. Including time as a predictor allows us to model the entire pitch contour of a tonal pattern by predicting F0 across the whole timespan of a token of that pattern. The GAM predicts F0 values not only for voiced sections but also for voiceless parts of the tokens by estimating what the pitch contour would likely be if the voiceless segments were voiced. In adopting this modeling strategy, our goal is not curve fitting, but cognitive modeling of pitch contours. We theorize that a speaker producing a word has a pitch contour for the whole word or even the whole utterance in mind. We further assume that the cognitive planning underlying the production of pitch contours is continuous, rather than a step function with jumps to zero Hz for voiceless segments. From our theoretical perspective, it is only at the stage of articulation that voiceless segments mask the internally projected pitch development.

The pitch contour predicted by the GAM, shown in the right-hand panel of Figure~\ref{fig:toy_contour}, captures the general trend with some precision, mirroring the raw data on the left. This graph is the model's best estimate of the average population contour for words with the RF tonal pattern, given the 12 tokens in our toy dataset. However, the empirical contours show considerable variation around this average, even for a single speaker producing citation forms in isolation. This variability in realization is the focus of our study.

To investigate whether individual tonal realizations might be related to the meaning of the carrier word token, we enriched the model with a by-word factor smooth, which is effectively a nonlinear, time-dependent, random effect for word type. In the present example, provided purely for illustration of the method, the by-word smooths are based on just two tokens of each word.\footnote{For detailed discussion of the ways in which these smooths can be specified, and the accuracy of these smooths, see \citet{baayen2022note}.} This mixed model predicts, for each word type, a word-specific adjustment contour which has to be added to the population pitch contour to obtain the predicted pitch for a given word type. 

\begin{figure}
  \centering
  \includegraphics[width=0.36\textwidth]{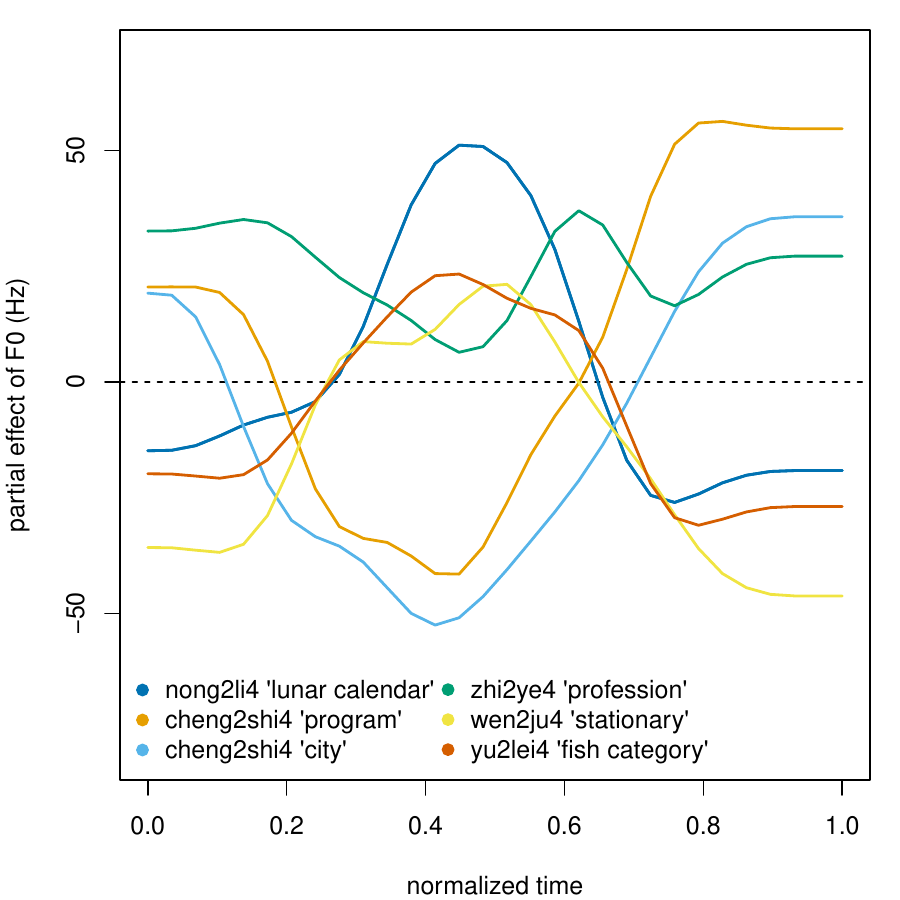}
  \hspace{-0.1cm}
  \includegraphics[width=0.57\textwidth]{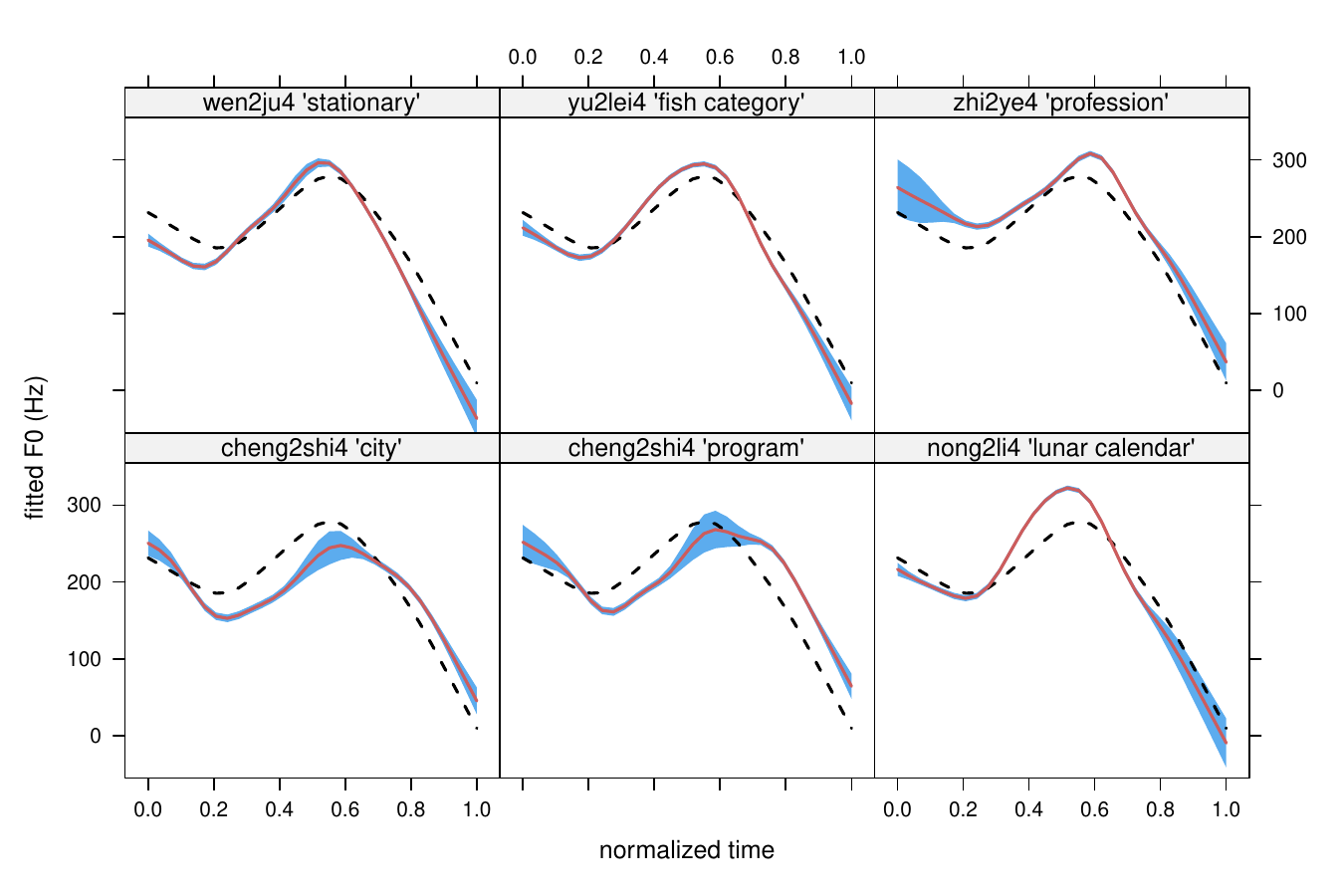}
  \caption{The left-hand panel shows by-word adjustment contours from the toy model with only by-word factor smooth and normalized time as predictors. The right-hand panel plots the fitted contour for each word, with the predicted general contour (identical for all words) indicated by the dashed line.}
  \label{fig:toy_fs}
\end{figure}

The by-word adjustment contours estimated by the GAM are visualized in the left-hand panel of Figure~\ref{fig:toy_fs}. The dotted line at $y=0$ is a reference line: an adjustment curve for a given word that followed this line would indicate that no adjustment is needed and that this word's pitch is identical to the population contour. Deviations above this reference line indicate an upward F0 adjustment, and deviations below it indicate a downward adjustment. The word \mandarin{職業} {\em zhi2ye4} `profession', for example, represented by a green curve, requires an upward adjustment for the entire contour, although the amount of adjustment varies across time. When a given word's adjustment contour is added to the general contour, we obtain its fitted contour, as shown in the right-hand panel of Figure~\ref{fig:toy_fs}. The dashed line in each graph represents the general contour which is, by definition, the same for all words. The red line (along with its confidence interval in blue) plots the fitted contour for the word in question. These fitted contours vary from word to word. For \mandarin{職業} {\em zhi2ye4} `profession', presented in the right-most upper panel, the entire fitted contour is above the general average contour, as expected. The homophone pair \mandarin{城市} {\em cheng2shi4} `city' and \mandarin{程式} {\em cheng2shi4} `computer program', shown in the left and middle lower panels, have similar but not identical fitted contours, as would be expected if word meaning, as well as word form, plays a role in determining tonal realization.

Because the GAM receives no input for voiceless sections of the tokens, where there is no actual pitch, the model's estimations for these sections are more uncertain than for the voiced sections. Similarly, word-specific partial effects for voiceless sections tend towards zero, i.e., no effect, again because there is no data. For instance, in Figure \ref{fig:toy_fs}, the panel for \mandarin{程式} {\em cheng2shi4} `program' shows that the GAM produces a wide confidence interval for the two voiceless consonants, exactly as it should. Because there is no data for the voiceless segments of a given token, the model falls back on the overall smooth of time in these intervals, i.e., the general rise-fall tonal pattern. Hence, in Figure \ref{fig:toy_fs}, it can be seen that the wide confidence intervals for voiceless sections of the tokens tend towards including the dotted line that represents the average contour.

Although this is just a toy example, it does illustrate two important aspects of the more detailed analyses reported below. Firstly, we can decompose the observed pitch contour of any token into a general population contour plus various more specific contours, including a meaning-specific adjustment contour. Secondly, GAMs can identify such meaning-specific contours and, given an adequate sample size, they could inform us about whether including meaning-specific contours improves model fit. 

In what follows, we turn to a much larger dataset of spontaneous spoken Taiwan Mandarin, and consider a much broader set of predictors that allow us to bring under statistical control a wide range of constraints known to co-determine the realization of pitch. If there is indeed a semantic component to the tonal realization of Mandarin disyllabic word tokens, then by-word factor smooths should be well-supported even when relevant control variables are taken into account.

\subsection{Data}\label{sec:data}
 
We used the Taiwan Mandarin spontaneous speech corpus, which is one of a pair of corpora created by researchers at the National Taiwan University \citep[see][for a description of the general method in the context of the Southern Min corpus]{fon2004}. The Mandarin corpus consists of about 30 hours of recorded interviews with 55 native Taiwan Mandarin speakers aged between 20 and 60 years, 31 self-identified as female and 24 self-identified as male, recruited through snowball sampling starting with friends, acquaintances or relatives of the team. The recordings were made over several years between 2000 and 2010, using high-quality microphones and digital recorders in quiet locations, whenever possible in a soundproof laboratory. Before the interview, participants were told that the purpose was to understand their views on changes in various aspects of life. The interviewers aimed not to engage in conversation with participants but to elicit longer monologues about their personal experiences in childhood, school, work, relationships and elsewhere. After the interview, participants were debriefed and asked for permission to include their recording in the corpus. If permission was granted, the researchers orthographically transcribed the recording using Chinese characters. The word segmentation system developed by Academia Sinica \citep{may2003introduction} was used to identify word boundaries in the orthographic representation, and dictionaries were used to add information about the canonical tones of the words identified. The character transcriptions were then romanized in order to run a forced aligner \citep[Easyalign:][]{goldman2011easyalign} to match the transcription to syllable boundaries in the audio files. The resulting alignment was manually checked and where necessary corrected.

The corpus described in the previous paragraph contains 94,783 disyllabic word tokens, 11,482 of which have the RF tonal pattern\footnote{These tokens also include tone sandhi words containing \mandarin{一} {\em yi1} and \mandarin{不} {\em bu4}, which have T2 realizations when followed by T4 syllables.}. These 11,482 tokens represent 707 orthographic word types. However, more than two-thirds of the tokens belong to one of five types: \mandarin{然後} {\em ran2hou4} `and then', \mandarin{時候} {\em shi2hou4} `during which', \mandarin{不會} {\em bu2hui4} `do not', \mandarin{還是} {\em hai2shi4} `still, or' and \mandarin{一樣} {\em yi2yang4} `likewise'. In order to avoid model predictions being heavily biased towards these five high-frequency words, we randomly sampled 300 tokens for each of these five types for inclusion in our dataset.\footnote{The figure of 300 was an arbitrary choice at the upper end of the frequency range for the other words in the dataset.} We also excluded types with fewer than 20 occurrences from the dataset, in order to avoid overfitting to low-frequency words. As a consequence, our initial dataset comprised 4,516 tokens across 53 word types.

We extracted the sound files for our 4,516 tokens and measured their F0 values using the method described in Section \ref{sec:methodology} for the toy dataset. For speakers identified as male, we set the pitch floor and ceiling to 50 and 400 Hz respectively. In order to make sure every token had sufficient data points for model fitting, we excluded extremely short tokens with fewer than six data points, which constituted about 5\% of the data. Next, we removed tokens where an F0 extraction error was likely. F0 extraction errors usually result from pitch halving or doubling, and lead to abrupt big changes in the recorded F0 values. We therefore first obtained, for each token, all the F0 differences between consecutive measurements, and then calculated the standard deviation of these difference values. The standard deviation is large when F0 measurements are discontinuous and fluctuate abruptly. Tokens with standard deviations greater than the 9th decile of the distribution were considered to be outliers, hence likely to involve extraction errors, and were excluded from further analyses. Finally, two words were excluded because their tokens were all contributed by only one speaker. 

The final dataset for the first analysis, reported below in Section \ref{sec:word_id}, consists of a total of 3,778 tokens representing 51 word types. Since these types do not include any heterographic homophones, there is a one-to-one correspondence between the orthographic labels of the tokens in our data and their canonical spoken forms. We therefore assume that tokens with the same label bear some similarity to one another in both form and meaning, i.e. belong to the same word type. However, because these tokens were extracted from spontaneous speech, not every speaker produced every type. To ensure that no word was completely nested under speaker or vice versa, we checked that every type was produced by multiple speakers and that every speaker produced multiple types. In the dataset of 3,778 tokens, the median number of speakers per word type is 20, the mean is 24.45, and the range is 5--52.  The median number of word types per speaker is 23, the mean is 22.67, and the range is 12--35.

In the dataset described above, the F0 values are positively skewed. We therefore log-transformed them to create a response variable with a distribution closer to Gaussian, a requirement for modeling with Gaussian GAMs.\footnote{For studies of auditory comprehension, several psychoacoustic scales are available, such as the Bark scale, which directly relate to human perception of pitch. However, as the interest of the present study is primarily in the pitch that speakers produced, we wanted to stay as close as possible to the physical signal and therefore simply log transformed the Hertz values.} Note that despite voiceless gaps in the data for certain tokens, the overall distribution of data points from all tokens combined is dense right across the [0,1] interval of normalized time. The model is therefore able to make accurate predictions for all tokens within this range, including those with voiceless segments, albeit with variations in confidence as described for the toy dataset in Section \ref{sec:methodology}.

\subsection{Predictors}\label{sec:predictors}

The core predictors in our GAM models are described in Section \ref{sec:interest} below. As far as possible, we also included as controls all the variables that have previously been shown to influence tonal realization, as outlined in Section \ref{sec:introduction} above. These control predictors can be grouped into three major categories: speaker-related, context-related, and segment-related. They are described in Sections \ref{sec:speaker} to \ref{sec:form} respectively. 

\subsubsection{Core predictors}\label{sec:interest}

\noindent
\textit{Word type:} We coded each word token in our dataset for its word type (\texttt{word}), using the orthographic representation of the token in the corpus as the identifier of its word type.\footnote{Note that, although word frequency is an important predictor for response variables such as spoken word duration, it is not a factor that has been widely reported to co-determine the shape of pitch contours in Mandarin \citep[but see][]{bi2015effect}. We verified for our data that when word frequency is included in a model that also has access to word type, frequency is not significant, whereas word type is well-supported. In this study, frequency of use is therefore not discussed any further. 
}

\vspace{0.5\baselineskip}
\noindent
\textit{Sense:} Unlike heterographic homophony, homographic homophony and polysemy are common in Mandarin disyllabic words. In lexicography, such diversity of meaning is usually addressed by attempting to enumerate the various possible senses of a given orthographic form. Similarly, in computational semantics, systems have been devised for disambiguating word senses from amongst a finite set of possibilities. The validity of this approach has been questioned, e.g. by \citet[p.~29]{kilgarriff2006word}, who pointed out that there are `no decisive ways of identifying where one sense of a word ends and the next begins'; polysemy is actually much more subtle and nuanced than a set of discrete possibilities would suggest. Nevertheless, sense annotations do capture, however crudely, some aspects of the variability in words' meanings. Furthermore, within the context of modeling pitch contours with GAMs, discrete senses are convenient because we can straightforwardly estimate specific pitch contours for each sense. We therefore coded every word token in the dataset for \texttt{sense}, using a word sense disambiguation system \citep{rocling2020} based on the Chinese Wordnet \citep{Huang2010}. The possible values of this variable correspond to the senses identified by the disambiguation system. More than one sense was identified for 35 of the 51 words in the dataset, with a total of 130 senses overall. All except two of the words had between one and five senses; of the two outliers, one had 6 senses and the other had 9 senses. Note that, because the sense labels in our data are nested under the orthographic form, and there are no synonyms, \texttt{sense} includes all the information in \texttt{word}, plus additional information about the meaning of any given token.

\vspace{0.5\baselineskip}
\noindent
\textit{Normalized time:} The points in time at which F0 measurements were taken were, for each token, transformed into a normalized time scale of 0-1 to produce the variable \texttt{time}.

\subsubsection{Speaker-related controls} \label{sec:speaker}
\noindent
\textit{Gender:} Speakers identified as female usually have a higher pitch register and wider pitch range than speakers identified as male. Furthermore, with respect to tonal realizations in Taiwan Mandarin, a number of studies have documented detailed gender-dependent differences in various sociolinguistic domains \citep{fu1999chinese, wu2003sociolinguistic, huang2008dialectal, wu2009effect}. We therefore included \texttt{gender} as a simple control variable to account for intrinsic pitch height and range differences between speakers of different genders, as labeled in the corpus, and also allowed \texttt{gender} to interact with \texttt{time}, to accommodate possible gender-specific modulations of the pitch contour. 

\vspace{0.5\baselineskip}
\noindent
\textit{Speaker identity:} Speaker identity was included to account for any idiosyncratic tonal realizations specific to individual speakers. We included \texttt{speaker} not only as a main effect, but also in interaction with \texttt{time}, using by-speaker factor smooths. 

\subsubsection{Context-related controls} \label{sec:context}

\noindent
\textit{Duration:}  The shapes of tonal contours are influenced by the time available to articulate them. Under time pressure, i.e., when words are spoken quickly, speakers do not have enough time to realize the full tonal contours, which are therefore more likely to deviate from their underlying shapes \citep{cheng2015mechanism, tang2020acoustic}. We measured the duration of each token in seconds. As the distribution of these measurements was heavily skewed to the right, we log-transformed them before conducting GAM analyses. In our models, the variable \texttt{duration} is the log-transformed token duration.\footnote{In our data, token duration is strongly correlated with speech rate (r =-0.57, p < 0.0001). In other words, these two variables could not meaningfully be included as predictors in the same model (cf. Section \ref{sec:strat}). Since duration turned out to predict F0 better than speech rate did, we report models with duration.}

\vspace{0.5\baselineskip}
\noindent
\textit{Adjacent tones:} When a tone is expected to start at a different pitch from where the previous one ends, e.g., a falling tone followed by a high level tone, the degree of co-articulation, and hence deviation from the canonical tonal shapes, will be greater than when two tones are contiguous, e.g., a high level tone followed by a falling tone \citep{xu1994production, shih1988tone}. In addition, although the details differ across studies, tonal co-articulation is usually found to be bi-directional, i.e., both anticipatory and preservatory \citep{shen1990tonal, xu1997contextual, huang4637487production}. For our analyses, we therefore coded the tonal category of each token's preceding and following syllables in the corpus. When a target token occurred utterance-initially or utterance-finally, the preceding or following tonal category was coded as `null'. This gave us six possible tonal categories for both the preceding syllable and the following syllable: four lexical tones, one neutral tone, and `null'. We therefore created the factor \texttt{adjacent\_tone} with 36 levels to account for each possible combination of properties of the preceding and following syllables. 

\vspace{0.5\baselineskip}
\noindent
\textit{Utterance position:} The realization of tone in an utterance is also influenced by sentence intonation \citep{ho1976acoustic, tseng1981acoustic, shen1989interplay, shen1990prosody}. For example, statement intonation is often characterized by a downward trend, resulting in pitch declination \citep{shih1997declination}. Question intonation, on the other hand, can potentially lead to a final rise, although this largely depends on the syntactic structure and/or emotive force of the question concerned \citep{lee2005prosody, chuang2007icphs}. For the current study, we simply calculated the normalized position of each token in the relevant utterance. We defined an utterance as a sequence of words preceded and followed by a perceivable pause (regardless of duration), as indicated by the labels provided in the corpus. The variable \texttt{utterance\_position} is the position at which a given token occurs in an utterance divided by the total number of words in that utterance. This predictor is therefore bounded between 0 and 1. For utterances with only one word, the utterance position was set to 1.\footnote{In many languages, F0 at the start of an utterance is related to the length of the utterance, with higher initial F0 for longer utterances. We therefore tried including utterance length in our models. Although overall model fit improved, the concurvity score of utterance length was as high as 0.84, suggesting that a large portion of the effect can be explained by other factors in the model (cf. Section \ref{sec:eval_word}). In particular, utterance length is negatively correlated with token duration (r =-0.23), so keeping both variables would have rendered their effects uninterpretable. We chose to remove utterance length and keep duration.}

\vspace{0.5\baselineskip}
\noindent
\textit{Bigram probability:} Bigram probability is a measure of a word's contextual predictability based on its relative frequency of co-occurrence with the other words in its context; the higher the bigram probability, the more predictable a target word is in the given context. It has been found that a word's phonetic realizations are intimately related to its contextual predictability. In general, higher predictability is associated with shorter word duration and a greater degree of spectral reduction \citep{bell2003effects, gahl2012reduce}. Specifically for tonal realizations in Mandarin, there is some evidence that these too are sensitive to contextual predictability; when a word is more contextually predictable or represents given information, its F0 excursion and range are found to be diminished \citep{hsieh2013prosodic,ouyang2015prosody}. In the present study, following \citet{gahl2012reduce}, we calculated the bigram probabilities of target tokens in two ways: \texttt{bigram\_previous}, based on the preceding word, and \texttt{bigram\_following}, based on the following word. These two variables are defined, respectively, as follows: 
\begin{align*}
  P(w_n|w_{n-1}) = \mbox{Freq}(w_{n-1}, w_n)/\mbox{Freq}(w_{n-1}),  \\   P(w_n|w_{n+1}) = \mbox{Freq}(w_n, w_{n+1})/\mbox{Freq}(w_{n+1}),
\end{align*}
where \(P(w_n|w_{n-1})\) is the probability of a word occurring given the previous word, \(P(w_n|w_{n+1})\) is the probability of a word occurring given the following word, and Freq denotes word frequency in the corpus of Taiwan Mandarin.

\subsubsection{Segment-related controls}\label{sec:form}

\noindent
\textit{Vowel height:} It has long been recognized that different vowels have different intrinsic pitch, a finding established for a great number of different languages, including Mandarin \citep{whalen1995universality, ladd1984vowel, ho1976acoustic, shi1987vowel}. Specifically, high vowels tend to have higher F0 values than low vowels. For our disyllabic words, we coded the vowel heights of the vowels of the first and second syllables as two separate predictors, \texttt{vowel1} and \texttt{vowel2} respectively. For monophthongs such as /i/ and /a/, we distinguished between three vowel heights: `high', `mid', and `low'. For diphthongs such as /a\textipa{I}/ and /e\textipa{I}/, which are characterized by within-vowel changes in height, we added two additional levels: `low-high' and `mid-high'. This means that there are theoretically 25 possible combinations of \texttt{vowel1} and \texttt{vowel2}. Our dataset included 20 of these possible combinations.

\vspace{0.5\baselineskip}
\noindent
\textit{Onset:}
The effect of onset consonant on F0 has been studied in considerable detail in Mandarin. \citet{ho1976acoustic}, for example, found that after voiced consonants, pitch tends to start lower than after voiceless consonants. For stops, aspiration also results in lower initial F0, although the magnitude of this effect appears to be tone dependent \citep{xu2003effects}. Following \citet{howie1974domain}, we distinguished onset types according to manner of articulation, voicing, and aspiration. For each of the two syllables in our target words, we distinguished between `aspirated-affricate', `aspirated-stop', `unaspirated-affricate', `unaspirated-stop', `voiceless-fricative', and `voiced'. Syllables that do not have onsets and start with vowels or glides instead were coded as `null'. Our dataset contained 30 different combinations of the onset type of the first syllable (\texttt{onset1}) and that of the second syllable (\texttt{onset2}).

\vspace{0.5\baselineskip}
\noindent
\textit{Rhyme structure:} Although effects appear to be unstable and are not always reliably observed, some studies have reported variation in F0 for different Mandarin syllable types \citep{howie1974domain, xu1998consistency, fon2007positional}. In our models, we therefore included a control variable for syllable structure. Given the strict phonotactic constraints governing the syllables of Mandarin, a syllable can maximally be composed of an onset consonant, a prenuclear glide, a nucleus vowel, and finally a coda consonant \citep{duanmu2007phonology}. In some theoretical descriptions, a coda consonant must be a nasal; in other descriptions, it can be either a nasal or a postnuclear glide as in the case of /\textipa{aI}/, for example. In this study, we coded the latter cases as diphthongs in the vowel height predictors, \texttt{vowel1} and \texttt{vowel2}, and only coded for a coda consonant when the syllable included a final nasal. For each of the two syllables in our target words, we therefore coded the structure of the rhyme as `V', `GV', `VN', or `GVN'. This coding specifies, for a given syllable, whether there is a prenuclear glide, as well as whether there is a final nasal. Applied to the two syllables of our target words separately (\texttt{syllable1} and \texttt{syllable2}), we obtained 14 attested combinations of rhyme structures.

\subsection{Modeling Strategy}\label{sec:strat}
Because pitch typically changes continuously and gradually across the time course of an utterance, it is inevitable that our response variable of F0 measurements is characterized by significant autocorrelation. That is, the F0 at time $t$ is correlated with and can thus to some extent be predicted from the F0 at $t-1$. Autocorrelation is particularly problematic for regression modeling because the residuals of an autocorrelated response variable are also unavoidably autocorrelated. This means that a central assumption of regression modeling is violated, namely that the residuals should be independent of one another. 

In GAMs, the issue of autocorrelation can be addressed by incorporating a first order autoregression model for the errors, denoted as \textbf{AR(1)}. An AR(1) model is a linear model that predicts a given value of a time series from the immediately prior value; including an AR(1) process in a GAM enables the model to accommodate structure in the residuals by positing a linear relationship between a given residual and its preceding residual. Specifically, the residual at time $t$ is modeled as the sum of a proportion $\rho$ of the residual at time $t-1$ plus Gaussian noise. In all the models presented below, the AR(1) process is included. In order to determine the appropriate value of $\rho$, we first fitted a model without AR(1), then calculated the autocorrelation of the residuals in this model at lag 1, i.e., comparing the residual at each time point with the immediately preceding residual. Finally we fitted the model again now with AR(1), for which we set $\rho$ to the lag 1 autocorrelation.\footnote{We note that the current implementation of AR(1) in the mgcv package is sub-optimal, as it only allows for one $\rho$ value per model, effectively treating the entire dataset as one big time series. However, the degree of autocorrelation actually varies considerably across the tokens in our dataset. Applying a single $\rho$ correction will therefore under-correct for some tokens but over-correct for others, and could thus create an unwanted effect on the distribution of the residuals. We discuss this further in the supplementary material.}

A standard method for evaluating the effects of different predictors in regression modeling is to compare nested models in which individual variables are progressively added or removed. However, this approach assumes that the predictors are relatively independent of one another, so that the contribution of each predictor to overall variance can be identified. If highly correlated predictors are entered into a single model, it becomes impossible to separate the effects of one from another and hence to assess their relative contributions. It is self-evident that the segmental makeup of a word cannot be independent of that word and that, in our models, the core predictor \texttt{word} is almost completely identified by the segment-related controls and vice versa. Thus, in a model that included both \texttt{word} and the controls, it would be impossible to evaluate the effects of the different predictors. The same applies to \texttt{word} and \texttt{sense}, since the latter is completely nested under the former. Consequently, the standard approach to model comparison was not available to us as a means of testing our predictions, since we were specifically aiming to tease apart the effects of different variables.

Instead of comparing nested models that would have been linguistically uninterpretable, we adopted the following two-fold modeling strategy. Firstly, in order to evaluate whether a given predictor was relevant for understanding F0 contours, we made use of Akaike's Information Criterion \citep[AIC,][]{akaike1998information}. AIC estimates the relative quality of alternative models for a given set of data, by balancing goodness of fit against the number of parameters for each model, penalizing excessively complex models that might overfit the data. The method can be used for non-nested models, and is therefore well suited to our data. The best model is the one with the lowest value of AIC.\footnote{For model comparison using AIC, the evidence ratio is calculated as $e^{\frac{\Delta AIC}{2}}$. A decrease of 10 AIC units, for example, means that the model with the smaller AIC is 148.4 times more likely than the alternative model to generate the observed data.} Secondly, we investigated the adequacy of our predictors by means of cross-validation. That is, we held out a small portion of our dataset as testing data, fitted our models to the remaining data, and then assessed model accuracy on the testing data. We repeated this process for 100 different data splits selected by stratified random sampling such that every word type was represented in both the training and test data, in similar proportions. In other words, the test sets consisted entirely of novel tokens but no novel types, simulating the situation for human language use with previous experience. Using cross-validation, we could directly assess the precision with which our models predicted novel, previously unseen, data, and establish whether inclusion of a predictor improved prediction accuracy.\footnote{The cross-validation approach we used is sometimes called repeated training/test split or Monte Carlo cross-validation. It is one of the most common cross-validation methods in machine learning \citep{zhang1993model, kuhn2013applied}.}

We used this modeling strategy to explore the first two predictions presented in Section \ref{sec:introduction}, repeated here for convenience:
\begin{enumerate}
  \item Word type will be a stronger predictor of tonal realization than all the previously established word-form related predictors combined.
  \item Information about a word's meaning in context will improve prediction of its tonal realization, compared with prediction based on word type alone.
\end{enumerate}

\noindent Prediction 1 is based on the hypothesis that the unique pitch contour of each spoken Mandarin word token is determined in part by the meaning of that token. If this is correct, then variations in tonal realization arise not only from the mechanical constraints on articulation previously described in the literature, but also as a result of form-meaning connections in the lexicon. We therefore investigated whether using \texttt{word} as a predictor would lead to a more precise model of tonal realization, as compared to a model using the set of segment-related predictors (e.g., vowel height) that have previously been identified as relevant. Since \texttt{word} incorporates not only the segmental form of a word type but also the associated semantics, our expectation was that it would be a superior predictor compared to all the previously identified segment-related variables considered jointly, when we controlled for contextual effects such as token duration and adjacent tones.

As a first step, we fitted a baseline GAM to logarithmically transformed F0, including all the speaker-related and context-related control variables described in Sections \ref{sec:speaker} and \ref{sec:context}, in interaction with \texttt{time}. We then fitted six further models, each with one segment-related control variable added to the baseline, allowing us to investigate the articulatory effects described in the literature. To compare these effects jointly against the effect of \texttt{word}, we fitted two additional models: one with all six segment-related control variables included in addition to the baseline, and the other with only \texttt{word} as an additional predictor. Both \texttt{word} and the segment-related controls (\texttt{vowel1}, \texttt{vowel2}, \texttt{onset1}, \texttt{onset2}, \texttt{syllable1}, \texttt{syllable2}) were modeled as factor smooths in interaction with \texttt{time}; in other words, all these variables have both random intercepts and random wiggly effects on the shapes of the predicted contours.

If Prediction 1 is correct, and we find an effect of word type, this will provide some support for the hypothesis that meaning contributes to the realization of tone. However, it will not enable us to draw any firm conclusion. This is because word type subsumes information about form as well as meaning. Even if \texttt{word} performs better than the combination of all previously identified segmental predictors, we cannot be certain that this difference is due to meaning; it is possible that some additional aspect of word form influences tonal realization but has not yet been identified as a predictor. However, at the word level, it is impossible to tease these effects apart. Since all our word types have the same RF tonal pattern, if we were to enter every detail of a token's segmental makeup into the model, we would effectively be specifying its word type. Prediction 2 addresses this issue by separating variation in meaning from variation in form.

Prediction 2 is based on the hypothesis that variation in tonal realization is partly determined by token-level variation in meaning, i.e., variation in meaning within word type. To address this prediction, we compared the model with a factor smooth for \texttt{word} as the sole predictor added to the baseline against a model in which a factor smooth for \texttt{sense} was the sole predictor added to the baseline, both in interaction with \texttt{time}. Recall that, because the sense labels in our data are nested under the orthographic form, \texttt{sense} includes all the information in \texttt{word}, plus additional information about the meaning of any given token. If semantics is indeed at issue, replacing \texttt{word} by \texttt{sense} should therefore improve model fit even further.

In the dataset for the word-type analysis, the frequency distribution of \texttt{sense} is skewed towards the right, with about half of the senses having no more than 13 tokens. To make sure that all senses included in the models had sufficient tokens for statistical evaluation, we therefore used only a subset of the data for the models used to investigate the effect of \texttt{sense}. Since the median number of tokens per sense in the dataset was 13.5, we excluded senses with fewer than 14 tokens. This left us with a dataset of 3,458 tokens representing 65 senses across a smaller set of 48 word types. We used this smaller dataset for models evaluating \texttt{sense} as a predictor.\footnote{For completeness, we note that 35 words in the smaller dataset have only one sense. Of these 35 words, six have only one sense listed in the Chinese Wordnet, and seven are not included in the vocabulary of the Chinese WordNet. For the rest, either the tagger only identified one sense, or only one sense of the word had more than 13 tokens in our dataset.} The statistical analysis proceeded in the same way as described above for word type, except that we added the additional model with a factor smooth for \texttt{sense} as the only predictor in addition to the baseline.

\subsection{The baseline GAM}\label{sec:baseline}
\begin{figure}
  \centering
  \includegraphics[width=0.325\textwidth]{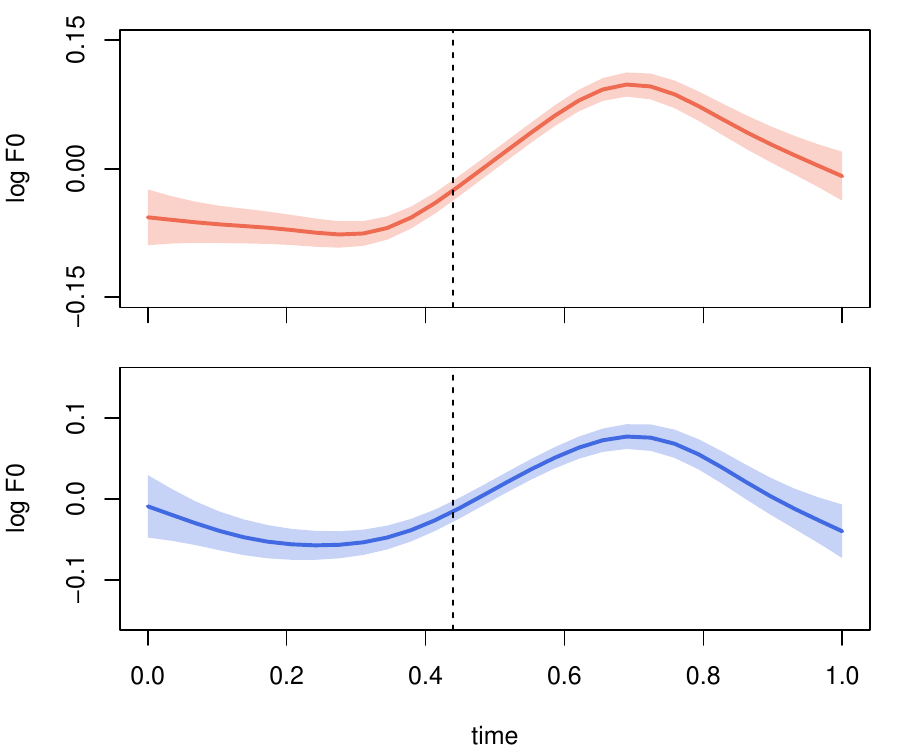}
  \includegraphics[width=0.325\textwidth]{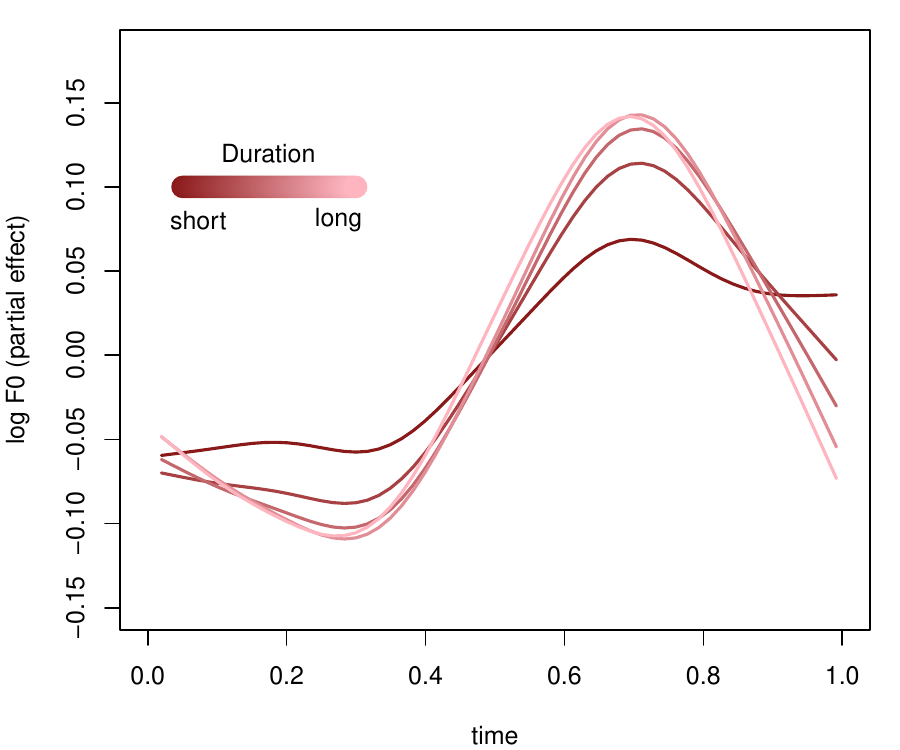}
  \includegraphics[width=0.325\textwidth]{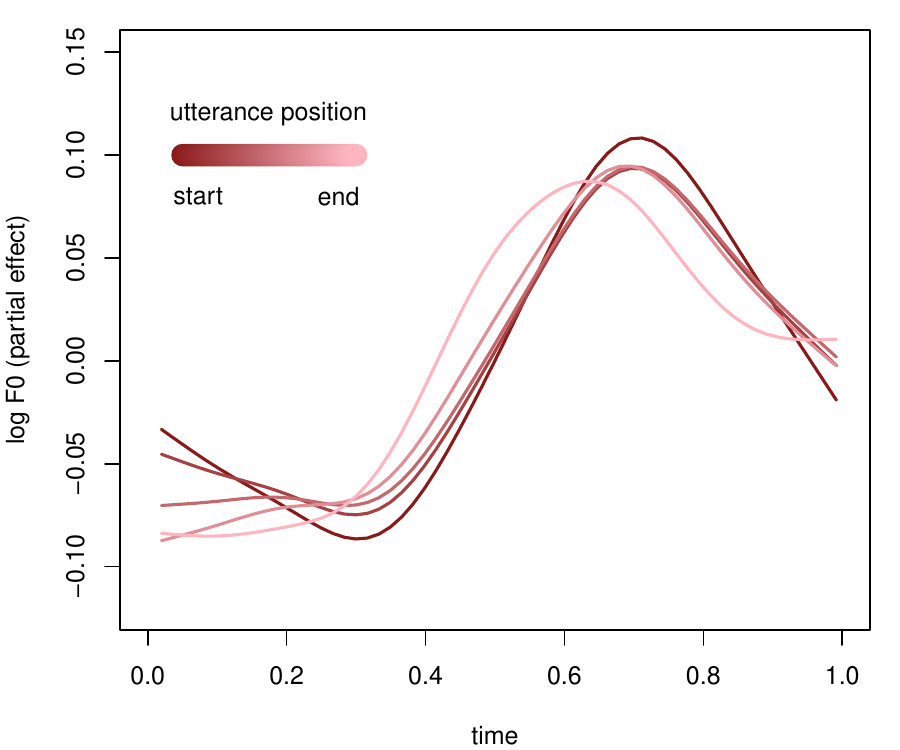}\\
  \includegraphics[width=0.325\textwidth]{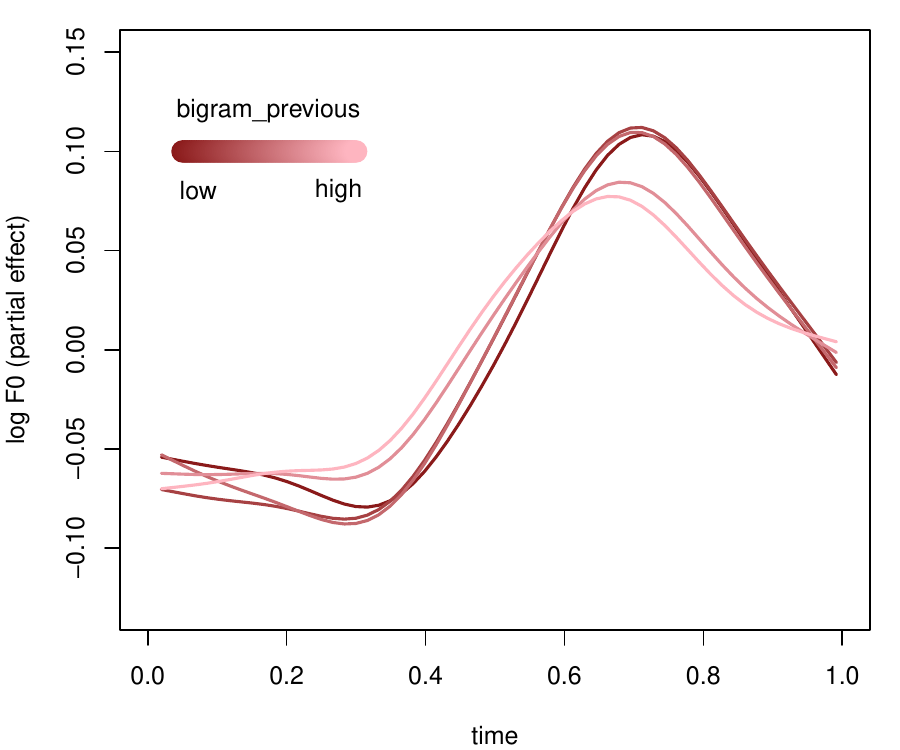}
  \includegraphics[width=0.325\textwidth]{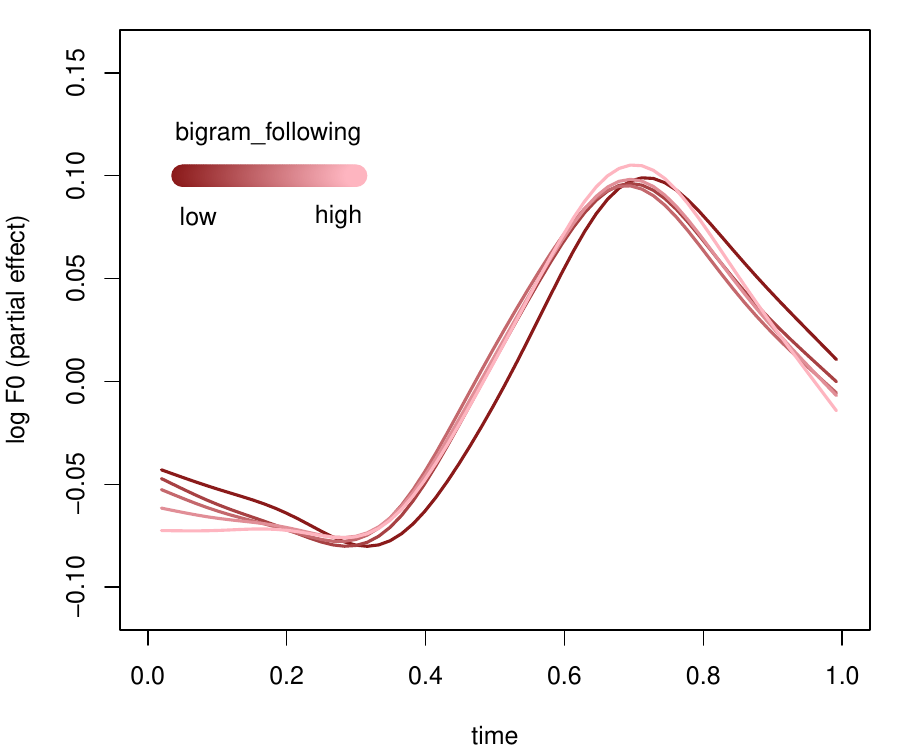}
  \includegraphics[width=0.325\textwidth]{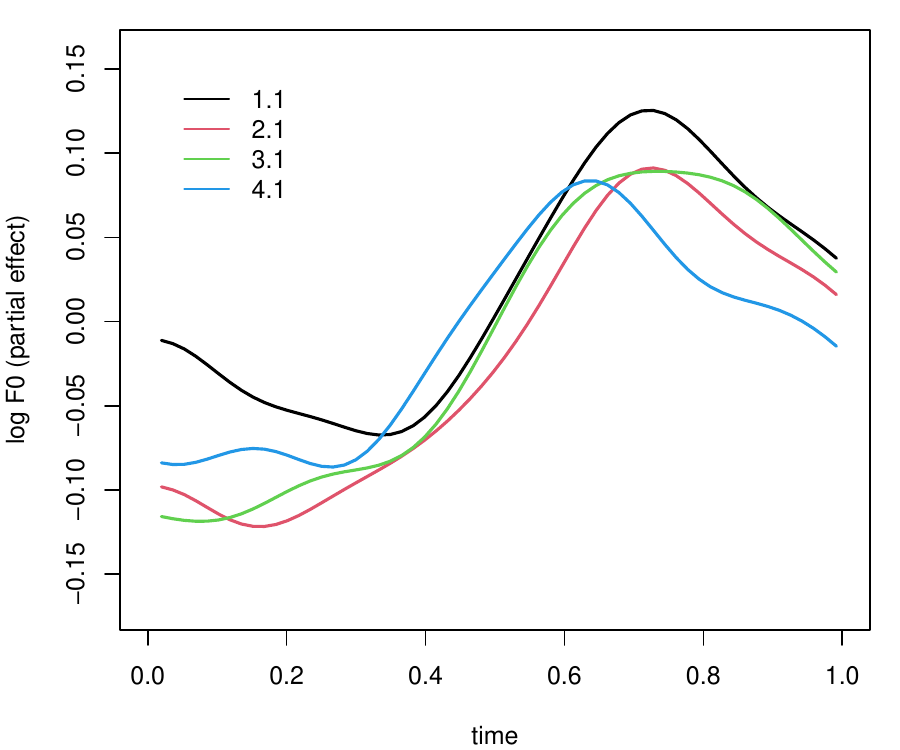}
  \caption{Partial effects in the baseline GAM. The upper left-hand panel shows the predicted base contours for speakers identified as female (red) and speakers identified as male (blue). The next four panels show, for female speakers, how the base contour is modulated by duration, utterance position, previous bigram probability, and following bigram probability, respectively. The final panel presents, again for female speakers, the effect of tonal coarticulation with the tone of the preceding word, when the following word has a high-level tone.}
  \label{fig:word_baseline_ef}
\end{figure}

We fitted the baseline GAM to log-transformed F0 (\texttt{pitch}), using the following model specification:\footnote{All model formulae and summaries are also provided in the supplementary material.} 
{\normalfont\ttfamily
\begin{tabbing}
xxxx\=xxmmxxxx\=\kill
  \> pitch $\sim$ \> gender + s(time, by = gender) + \\
  \> \> s(time, speaker, bs = 'fs', m = 1) + \\
  \> \> s(duration) + ti(time, duration) + \\ 
  \> \> s(utterance\_position) + ti(time, utterance\_position) + \\  
  \> \> s(bigram\_previous) + ti(time, bigram\_previous) + \\ 
  \> \> s(bigram\_following) + ti(time, bigram\_following) + \\
  \> \> s(time, adjacent\_tone, bs = 'fs', m = 1) 
\end{tabbing}
}
\noindent
The first line of this model requests a main effect for gender, in order to account for male voices being lower on average than female voices. We used treatment (i.e. dummy) coding, with female as the reference level. In addition, the model requests a separate smooth for each gender. The upper left-hand panel of Figure \ref{fig:word_baseline_ef} plots the predicted contours for speakers identified as female (red) and those identified as male (blue). Similar to the pattern observed in read speech (cf. Figure \ref{fig:toy_contour}), the realization of the RF tonal pattern in spontaneous speech is characterized by a shallow fall, followed by a long rise, and finally a much larger fall. Speakers identified as male show reduced pitch excursion compared to those identified as female, presumably due to male voices having a more compressed pitch range. The second line of the model requests by-speaker nonlinear random effects, using factor smooths. These factor smooths specify, for each speaker, the specific way in which that particular speaker modulates the general F0 contour associated with their gender.\footnote{The syntax \texttt{bs='fs'} on the second line of the baseline model specification has a similar effect to the \texttt{by} argument on the first line; both terms request a smooth for each level of a single factor variable (in this case, \texttt{speaker} and \texttt{gender} respectively). The \texttt{by} argument is generally preferred when the factor has few levels, and the levels are of interest, as in the case of \texttt{gender\texttt}. For computational efficiency, the \texttt{fs} argument is preferred when there are many levels and these levels are of less direct interest, as in the case of \texttt{speaker}.  When specifying a by-smooth, a separate term requesting a main effect for the intercept needs to be specified. 
In contrast, a factor smooth incorporates adjustments to the intercept, thus effectively calibrating the individual smooths for their relative position with respect to the general intercept. See \citet{baayen2022note} for detailed discussion.
}

The next four lines in the model formula deal with the four numerical context-related controls, namely \texttt{duration}, \texttt{utterance\_position}, \texttt{bigram\_previous} and \texttt{bigram\_following}. For each of these variables, the model requests a main effect smooth in combination with a tensor product smooth for the interaction of the given variable with \texttt{time} (using an interaction-specific tensor product smooth specified with \texttt{ti}). The upper mid panel of Figure \ref{fig:word_baseline_ef} plots the modulating effect of token duration on the base contour of speakers identified as female. Shorter duration, represented by darker shades of red, reduces the amplitude of the wave. The effect of position in the utterance is depicted in the upper right-hand panel of Figure~\ref{fig:word_baseline_ef}. The tonal shape is clearly most different when the word occurs towards the end of an utterance, in which case we observe an earlier peak. This might be due to the fact that we coded words in singleton utterances as occurring at the end of the utterance. The left and mid panels in the lower row of Figure~\ref{fig:word_baseline_ef} present the effects of the bigram probabilities given the preceding and following word respectively; higher bigram probabilities are represented by lighter shades of red. When \texttt{bigram\_previous} is high, meaning that the word is more expected given the preceding word, the F0 excursion is reduced. This effect parallels the finding of \citet{hsieh2013prosodic} that F0 excursion in Taiwan Mandarin is diminished in conditions of high semantic predictability. The effect of \texttt{bigram\_following} in our model is much smaller than the other contextual effects but appears to go in the opposite direction to \texttt{bigram\_previous}, with higher values associated with sharper peaks in F0.

The final line of the model specification requests factor smooths for \texttt{adjacent\_tone}, requesting a separate smooth for each of its 36 levels. The effect of \texttt{adjacent\_tone} is presented in the lower right-hand panel of Figure~\ref{fig:word_baseline_ef} for those tokens which have T1 as following tone. Unsurprisingly, the four predicted contours end similarly. However, the initial part of the contour diverges considerably, depending on the preceding tone. As expected, tonal context has a very large effect on the shape of the F0 contour. 

In what follows, we take this model as our baseline model, with all contextual covariates controlled for, and compare the effects of predictors representing individual aspects of word form with the effect of word type as a whole. 

\subsection{Results and discussion: word type}\label{sec:word_id}
\subsubsection{Evaluation of predictors}\label{sec:eval_word}
\begin{figure}
  \centering
  \includegraphics[width=.45\textwidth]{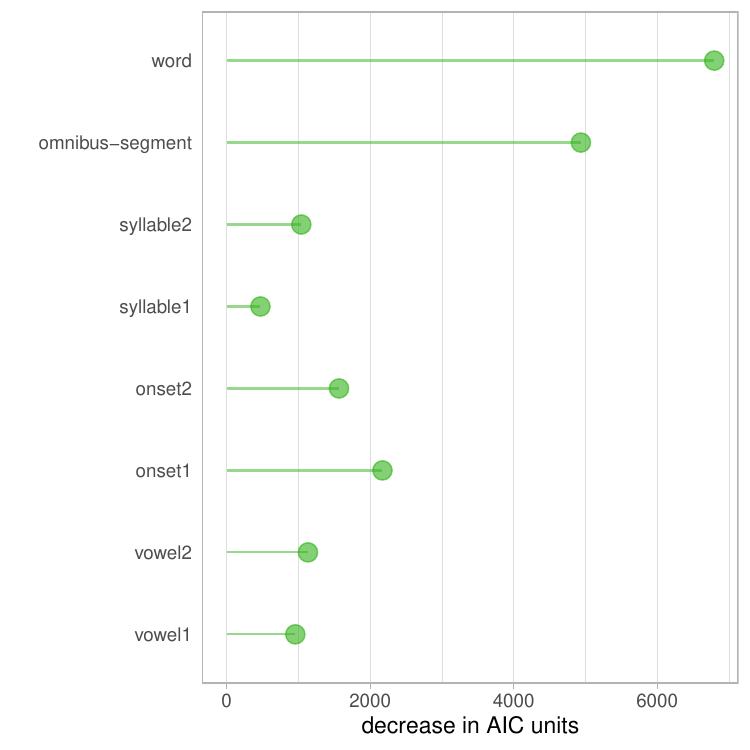}
  \includegraphics[width=.45\textwidth]{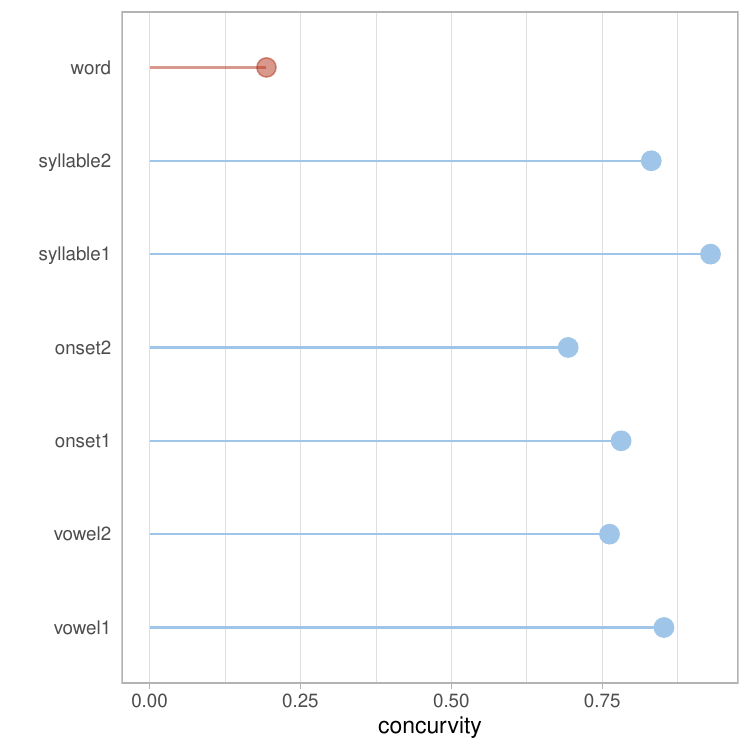}
  \caption{The left-hand panel shows model fit improvement gauged by decrease in AIC units when a predictor (or set of predictors) is added to the baseline model for the word-type analysis. The right-hand panel shows the concurvity score of individual predictors in two models using the full dataset of 3,778 tokens: the omnibus-segment model (blue) with factor smooths for all segment-related control variables added to the baseline, and the word model (red) with only a factor smooth for \texttt{word} added to the baseline.}
  \label{fig:comp_word_model}
\end{figure}

\noindent
The left-hand panel of Figure \ref{fig:comp_word_model} presents the relative improvement in model fit, as compared to the baseline model, for the six models with an additional factor smooth for a single segment-related control, the model with factor smooths for all segment-related controls (henceforth the \textbf{omnibus-segment} model), and the model with only a factor smooth for \texttt{word} as additional predictor. Improvement in model fit is gauged by the magnitude of any decrease in AIC. As can be seen, each individual segment-related control improves model fit, a result that dovetails well with the previous studies summarized in Section \ref{sec:introduction}. 

In the omnibus-segment model, the inclusion of all six segment-related controls jointly provides a substantially better fit than any single predictor and leads to a fall in AIC of 4,938 AIC units compared to the baseline model. However, this is a very poor model from a statistical point of view because its key predictors are correlated with one another. In a GAM, the \textbf{concurvity} score of a predictor is a number bounded between zero and one that measures the degree to which the effect of a given independent variable can be predicted by one or more of the other independent variables in the model.\footnote{Concurvity is the nonlinear equivalent of collinearity.} If a predictor's concurvity is low, this predictor has its own explanatory value; however, if the concurvity is high, the predictor is strongly confounded with other predictors. The blue dots in the right-hand panel of Figure \ref{fig:comp_word_model} represent the concurvity scores of the segment-related controls in the omnibus-segment model. It can be seen that the concurvity scores of all predictors are high, indicating that the effects of the segment-related controls are confounded with one another, rendering interpretation of the individual effects difficult if not impossible.\footnote{As explained in Section \ref{sec:strat}, this high level of concurvity, which would be further increased by the addition of \texttt{word}, is the reason why we cannot create a meaningful model that includes both \texttt{word} and the segment-related controls.} Despite the fact that the omnibus-segment model is linguistically uninterpretable, we have presented it here purely for comparison with the model with only \texttt{word} as additional predictor.

The concurvity score of the predictor \texttt{word} in the word-type model is represented by the red dot in the right-hand panel of Figure \ref{fig:comp_word_model}. It can be seen that \texttt{word} has low concurvity with the other predictors in this model (i.e., the baseline controls), so that the interpretation of the effect of individual word types on the F0 contours is straightforward. Furthermore, adding only {\texttt{word}} to the baseline model results in an even better fit than the omnibus-segment model, with a fall of 6,795 AIC units compared to the baseline. Although the segment-related controls address all the word-internal properties previously found to influence tonal realization, the contribution of just \texttt{word} by itself to the model fit is much stronger. The difference of 1,857 AIC units between the omnibus segment model and the word model means that the probability of the word model giving a better fit approaches infinity. It is clear that the association between word type and tonal realization cannot be reduced to the segment-related constraints on articulation previously described in the literature. The actual pitch contour is richer than what can be predicted from all phonetic features that have been found to be relevant. Note, however, that we do not claim that the segmental predictors are irrelevant, as our analyses clearly demonstrate that they all improve on the baseline model.  We also do not claim that once \texttt{word} is included, the segmental predictors are irrelevant.  What we do claim is that \texttt{word} by itself outperforms all segmental predictors jointly, and hence has added predictive value over and above the segmental predictors.

\begin{figure}
  \centering
  \includegraphics[width=0.8\textwidth]{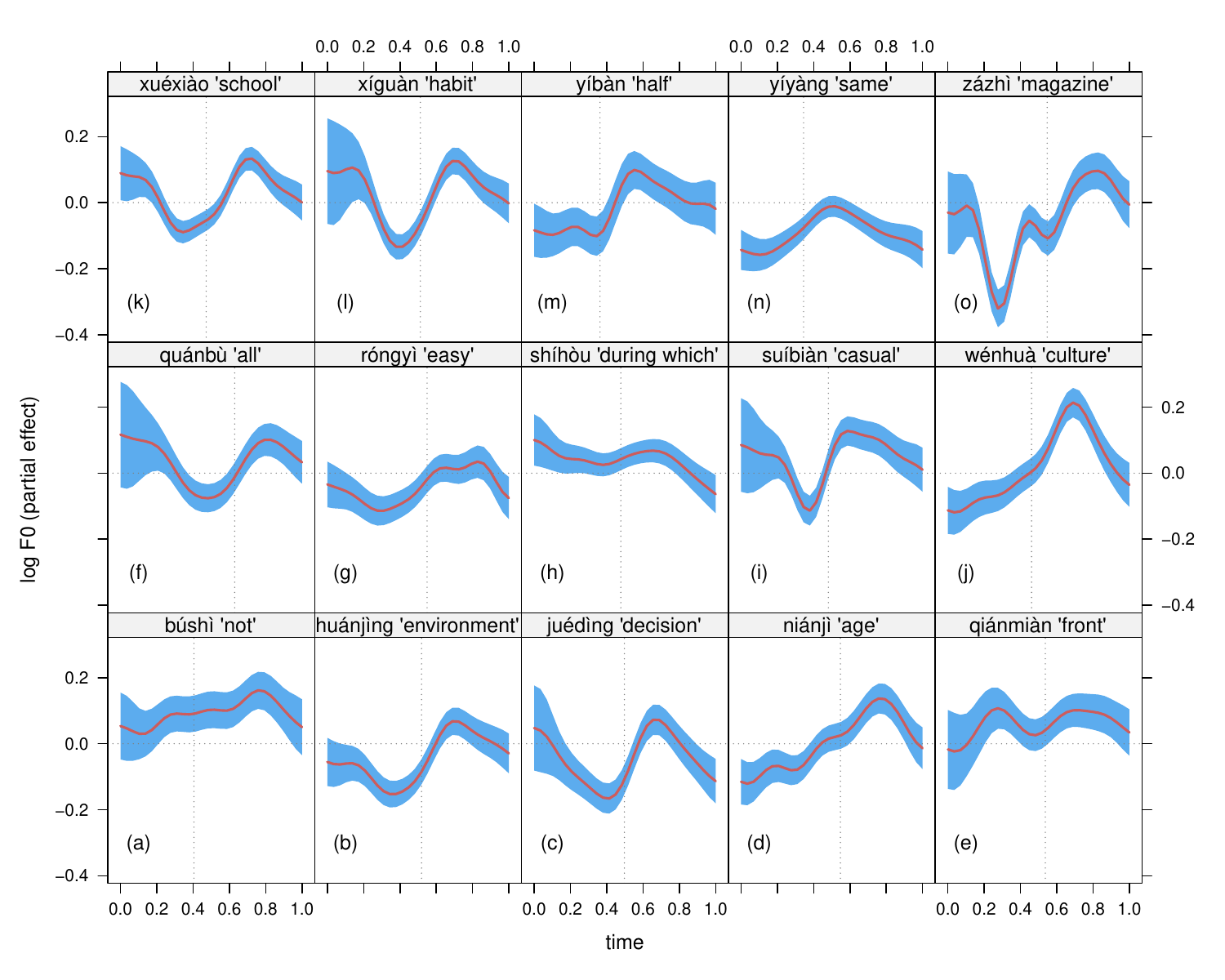}
  \caption{Examples of the pitch contours predicted by the general smooth for \texttt{time} for female speakers, combined with the partial effects of the factor smooth for \texttt{word}. These partial effects do not include the general intercept, nor the differences in pitch between female and male speakers. As they represent the pure effect of \texttt{word} on the pitch contour, irrespective of other predictors, the curves are centered around the y-axis (indicated by a horizontal dotted line). The vertical dotted lines in the panels indicate the average (word-specific) syllable boundary.}
  \label{fig:word_pred}
\end{figure}

The predicted pitch contours of a sample of 15 word types are presented in Figure \ref{fig:word_pred}. To better visualize how the word-specific tonal modulations differ from one another, the partial effect predicted for each word has been added to the general contour for speakers identified as female (cf. Figure \ref{fig:word_baseline_ef}). In general, the fall-rise-fall pattern can be observed for all these words, but the details of tonal excursions differ significantly from word to word. For example, while the initial falling part is very prominent for words like (c) \mandarin{決定} {\em jue2ding4} `decision' and (f) \mandarin{全部}, {\em quan2bu4} `all', it is rather muted for (m) \mandarin{一半} {\em yi2ban4} `half' and (d) \mandarin{年紀} {\em nian2ji4} `age'. In addition, in terms of the degree of undulation, some words have more reduced tonal range, such as (a) \mandarin{不是} {\em bu2shi4} `not' as compared to (j) \mandarin{文化} {\em wen2hua4} `culture', which has an extensive F0 excursion.

A closer inspection of Figure \ref{fig:word_pred} reveals that, as expected, some of the word-specific contours appear to be consistent with the words' canonical segmental properties. For example, the initial fall appears to be more salient when the onset of the first syllable is a voiceless sibilant, e.g., (l) \mandarin{習慣} {\em xi2guan4} `habit' or an affricate, e.g., (o) \mandarin{雜誌} {\em za2zhi4} `magazine', with the onsets /\textipa{\textctc}/ and /\textipa{ts}/ respectively. This pattern might be partly associated with the tendency for voiceless onsets to be followed by higher initial F0 in the following vowel, as compared to voiced onsets \citep{ho1976acoustic}. However, it should be kept in mind that the model is imputing F0 values for these voiceless onsets, where no periodic wave form is actually produced. In Figure \ref{fig:word_pred}, the plotted 95\% confidence intervals for the early timesteps in these words are wide, and can be seen to partially straddle the horizontal axis. 
Since the horizontal axis represents no effect, there is no good evidence for modulation of the general F0 contour early on in these words. Another segmental property that is to some extent visible in Figure~\ref{fig:word_pred} is the length of the second syllable, relative to the length of the first syllable. If the second syllable is relatively short as compared to the first syllable, e.g., (f) \mandarin{全部} {\em quan2bu4} `all' and (g) \mandarin{容易} {\em rong2yi4} `easy', the final fall tends to be attenuated, as expected given that relatively less time is available to physically implement a large fall in pitch. Nevertheless, the superior performance of \texttt{word} over the segment-related controls in our models suggests that such articulatory effects are not the only elements at play in determining tonal realization.

\subsubsection{Cross-validation}\label{sec:crossval_word}

Recall from Section \ref{sec:strat} that the goal of the cross-validation analysis is to investigate whether the fitted model can make precise predictions for held-out data, i.e. tokens that were withheld from the model during model fitting (training). If our hypothesis is correct, a GAM that has access to \texttt{word} should provide superior prediction accuracy on held-out data compared to a GAM that has access only to the segment-related controls. We therefore evaluated prediction accuracy under cross-validation. We held out 10\% of the current data as test data, and used the remaining 90\% as training data. 
Every word type was represented in both the training data and the test data, with approximately the same distribution in each set. We used stratified random sampling to produce 100 different splits with these properties.

We fitted ten models to the training data. In addition to the baseline model and the eight models assessed in Figure \ref{fig:comp_word_model} (left-hand panel), we added one more model that was given data in which the values of \texttt{word} were randomly permuted. That is, tokens of a given word were now assigned different random word labels. In what follows, we refer to this model as the \textbf{random-word} model. If the effect of \texttt{word} is genuine, then random permutation of the word labels should substantially reduce prediction accuracy. To quantify model accuracy, we obtained the models' predictions for the F0 contours of the held-out test data, and calculated the sum of squared errors (SSE) as a measure of prediction accuracy.\footnote{The sum of squared errors (SSE) is the sum of the squared difference between the observed and predicted values. A smaller SSE indicates more precise model predictions.} The SSE for the held-out data of a given model should be smaller than the SSE of the baseline model if the addition of one or more predictors indeed improves that model's prediction accuracy. We ran all ten models with all 100 random splits to cross-validate our results for the 100 held-out datasets.

\begin{figure}
  \centering
  \includegraphics[width=.48\textwidth]{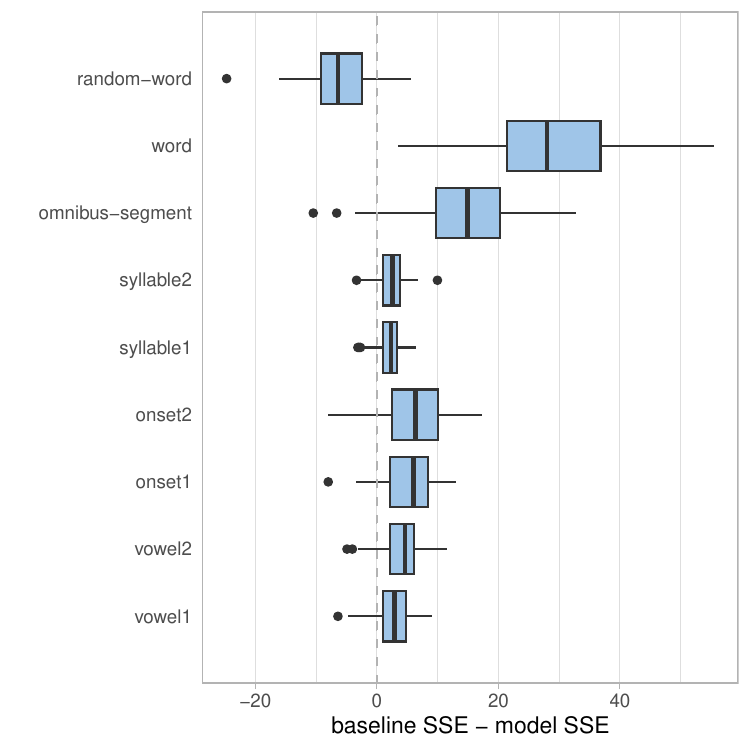}
  \caption{Model accuracy under 100 runs of cross-validation for the word-type analysis. The boxplots represent the distributions of reduction in SSE.}
  \label{fig:word_cv_res}
\end{figure}

Figure \ref{fig:word_cv_res} presents boxplots of the SSE difference between the baseline model and each of the nine models of interest.  
Positive values indicate that the model in question offers more precise predictions than the baseline, as its SSE is smaller than the baseline's. All the individual segment-related controls increase prediction accuracy over the baseline to some extent, albeit to varying degrees. However, the omnibus-segment model and the word model produce substantially greater increases in prediction accuracy, with the latter reducing the SSE to a larger extent than the former, replicating the model fit results. Moreover, when word labels are randomized, model accuracy plummets: the SSE of the random-word model is greater than that of the baseline model. 

The results presented so far provide strong evidence that word type is predictive of tonal realization, over and above the segment-related predictors established by previous studies. Our hypothesis, arising from the theoretical framework of the Discriminative Lexicon Model \citep[DLM][]{baayen2019discriminative,Chuang:Baayen:2021,Heitmeier:Chuang:Baayen:2024}, is that the predictive power of word type arises not only from articulatory constraints, but also from a close association between word meaning and phonetic form, which enables the language learner or user to discriminate more efficiently between forms with different meanings. However, as discussed in Section \ref{sec:strat}, the effect of word type cannot be unequivocally attributed to semantics, since in addition to word meaning, \texttt{word} captures all of a word's form properties, not only those previously identified as affecting tonal realization. Furthermore, since word meaning varies with context, there is no one-to-one correspondence between word type and the meaning of a given token; \texttt{word} can only encompass a rather general approximation of what each token means. To address both these issues, it is necessary to turn to Prediction 2 and the model with a factor smooth for \texttt{sense} as sole predictor added to the baseline. If word meaning is indeed predictive of tonal realization, then a model replacing the factor smooth for \texttt{word} by a factor smooth for \texttt{sense} should improve model fit even further.\footnote{Note, however, that as discussed in Section \ref{sec:predictors}, the senses that constitute the possible values of our \texttt{sense} variable discretize a much more subtle and interesting palette of shades of meanings.}

\subsection{Results and discussion: sense} \label{sec:sense_id}

\subsubsection{Evaluation of predictors}\label{sec:eval_sense}

\begin{figure}
  \centering
  \includegraphics[width=.45\textwidth]{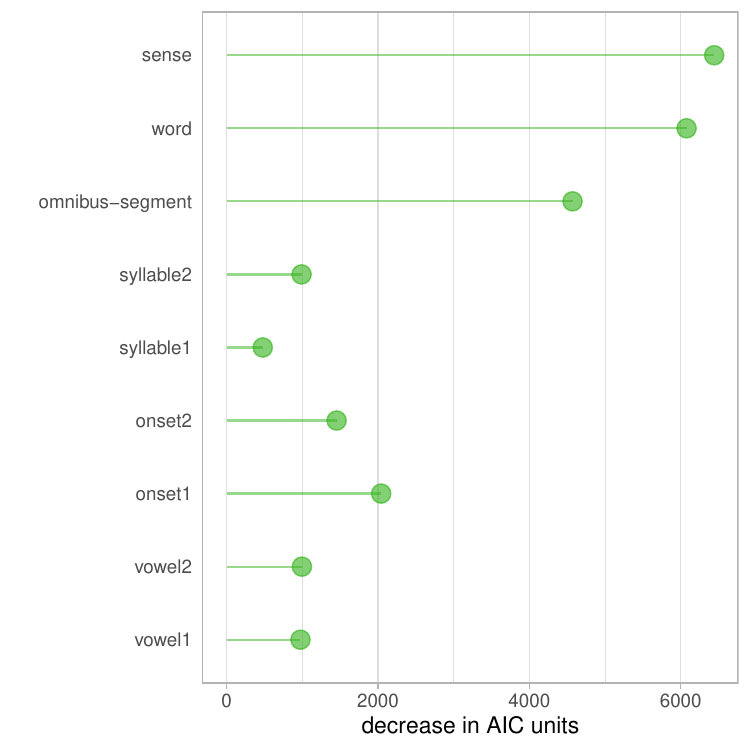}
  \includegraphics[width=.45\textwidth]{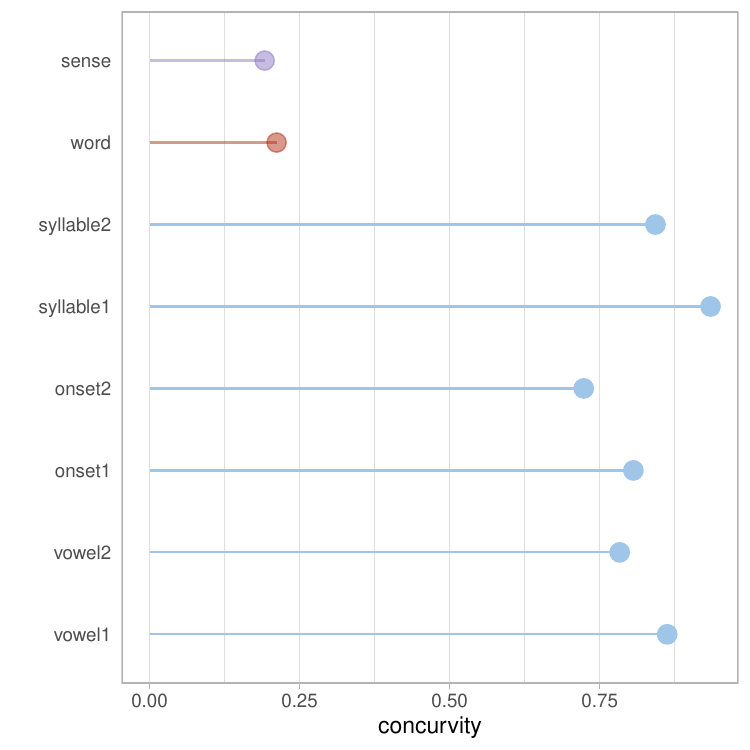}
  \caption{The left-hand panel shows model fit improvement gauged by decrease in AIC units when a predictor (or set of predictors) is added to the baseline model for the sense analysis. The right-hand panel shows the concurvity score of individual predictors in three models using the smaller dataset of 3,458 tokens: the omnibus-segment model with factor smooths for all segment-related control variables (blue), the word-type model with a factor smooth for \texttt{word} (red), and the sense model with a factor smooth for \texttt{sense} (purple).}
  \label{fig:comp_sense_model}
\end{figure}

\noindent
Figure \ref{fig:comp_sense_model} shows model fit improvement and concurvity for models based on the smaller dataset that included at least 14 tokens of each word sense.  Even with this smaller dataset, the overall pattern of results is very similar to that of the word type analysis. Critically, however, \texttt{sense} appears to be a somewhat better predictor than \texttt{word}. For this dataset, adding only {a factor smooth for \texttt{sense}} to the baseline model results in a fall of 6,443 AIC units compared with a fall of 6,078 AIC units when only a factor smooth for \texttt{word} is added. The difference of 365 AIC units means that the sense model is 1.81e+79 times more likely than the word model to explain the observed data. Furthermore, the effect of \texttt{sense} is also less confounded with the other predictors in the model (i.e., the baseline controls), as indicated by a smaller concurvity score.

\begin{figure}
  \centering
  \includegraphics[width=0.45\textwidth]{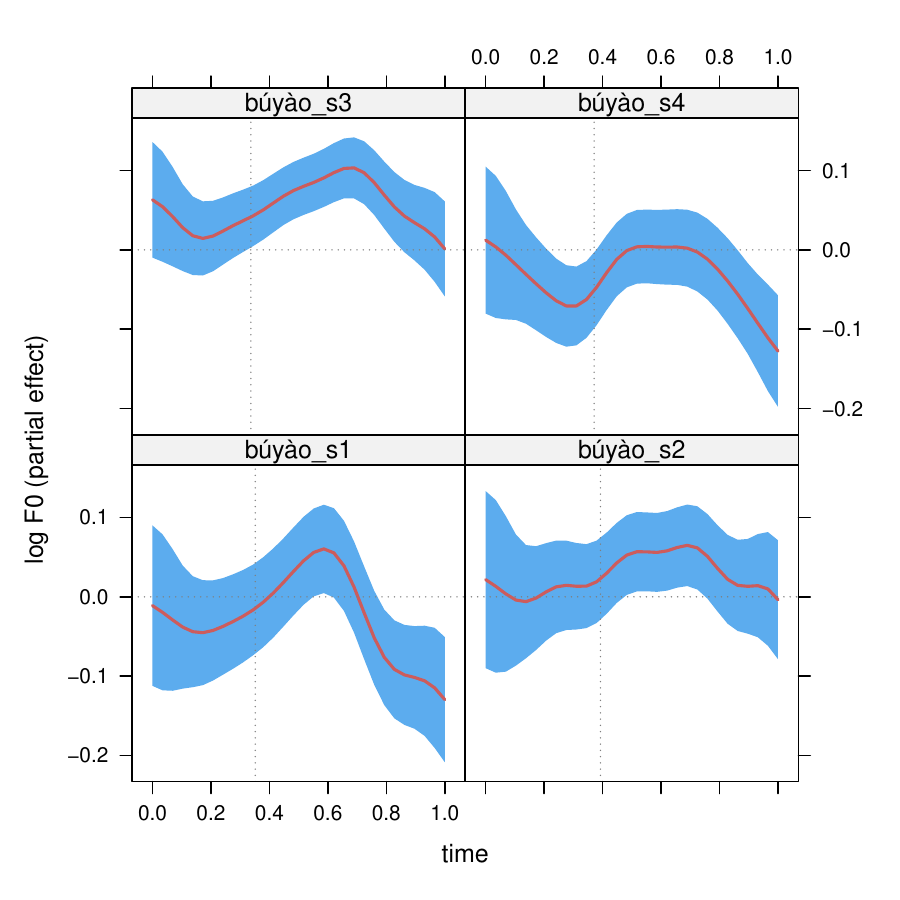}
  \includegraphics[width=0.4\textwidth]{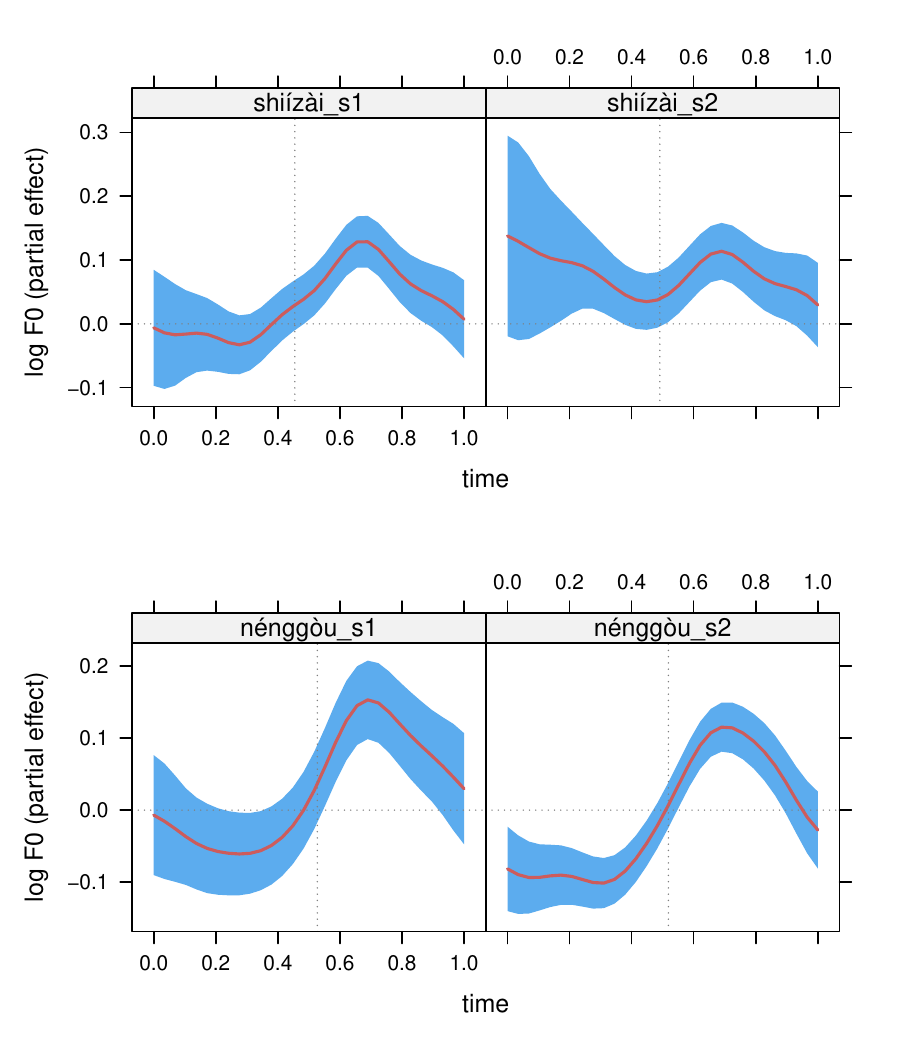}
  \caption{Examples of the pitch contours predicted by the general smooth
for \texttt{time} for female speakers, combined with the partial effects of the factor smooth for
\texttt{sense}. The left-hand panel shows the fitted tonal contours for different senses of the word \mandarin{不要} {\em bu2yao4}, a negation marker in Mandarin. The four senses are `prohibition', `dissuasion', `unneccesity', and `to wish something to not happen'. The upper right-hand panel shows the fitted tonal contours for the two senses of \mandarin{實在} {\em shi2zai4}, meaning `truly' and `indeed' respectively. The lower right-hand panel plots the fitted contours for the two senses of \mandarin{能夠} {\em neng2gou4}: `being capable of' and `enabling'.}
  \label{fig:sense_pred}
\end{figure}

Figure \ref{fig:sense_pred} presents the predicted tonal contours for different senses of three words: \mandarin{不要} {\em bu2yao4} (left), \mandarin{實在} {\em shi2zai4} (upper right), and \mandarin{能夠} {\em neng2gou4} (lower right). The word \mandarin{不要} {\em bu2yao4} is a polysemous negation marker in Mandarin. The four senses that are found in our dataset are `prohibition', `dissuasion', `unneccesity', and `to wish something to not happen' (s1 to s4, respectively). It can be seen that the different senses have clearly different tonal realizations. The panels on the right-hand side of Figure~\ref{fig:sense_pred} present the predicted contours for the other two words, each of which has two senses in our data. For \mandarin{實在} {\em shi2zai4}, tonal realizations vary greatly between the two senses (`truly' and `indeed'), whereas the realizations of the two senses of \mandarin{能夠} {\em neng2gou4} (`being capable of' and `enabling') are more alike, and differ mainly with respect to the amplitude of the pitch inflection. 

\subsubsection{Cross-validation}\label{sec:crossval_sense}
\begin{figure}
  \centering
  \includegraphics[width=.48\textwidth]{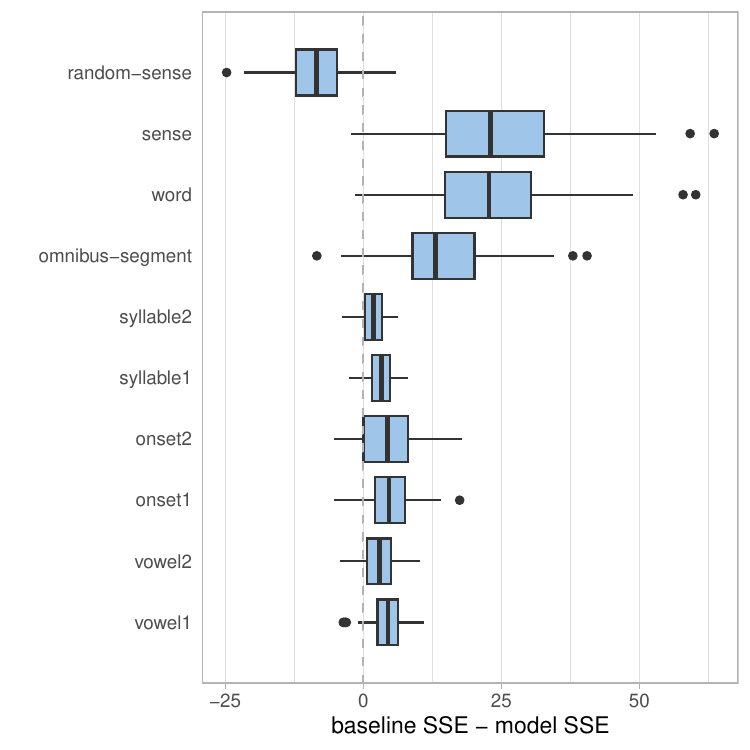}
  \caption{Model accuracy under 100 runs of cross-validation for the sense analysis. The boxplots represent the distributions of reduction in SSE.}
  \label{fig:sense_cv_res}
\end{figure}

As shown in Figure \ref{fig:sense_cv_res}, for the 100 cross-validation runs, it turns out that the model with \texttt{sense} is not necessarily always more accurate than the model with \texttt{word}. The medians of the reduction in SSE for the word and the sense model are very similar, with the variance of the sense model being somewhat larger. 

\begin{figure}
  \centering
  \includegraphics[width=.8\textwidth]{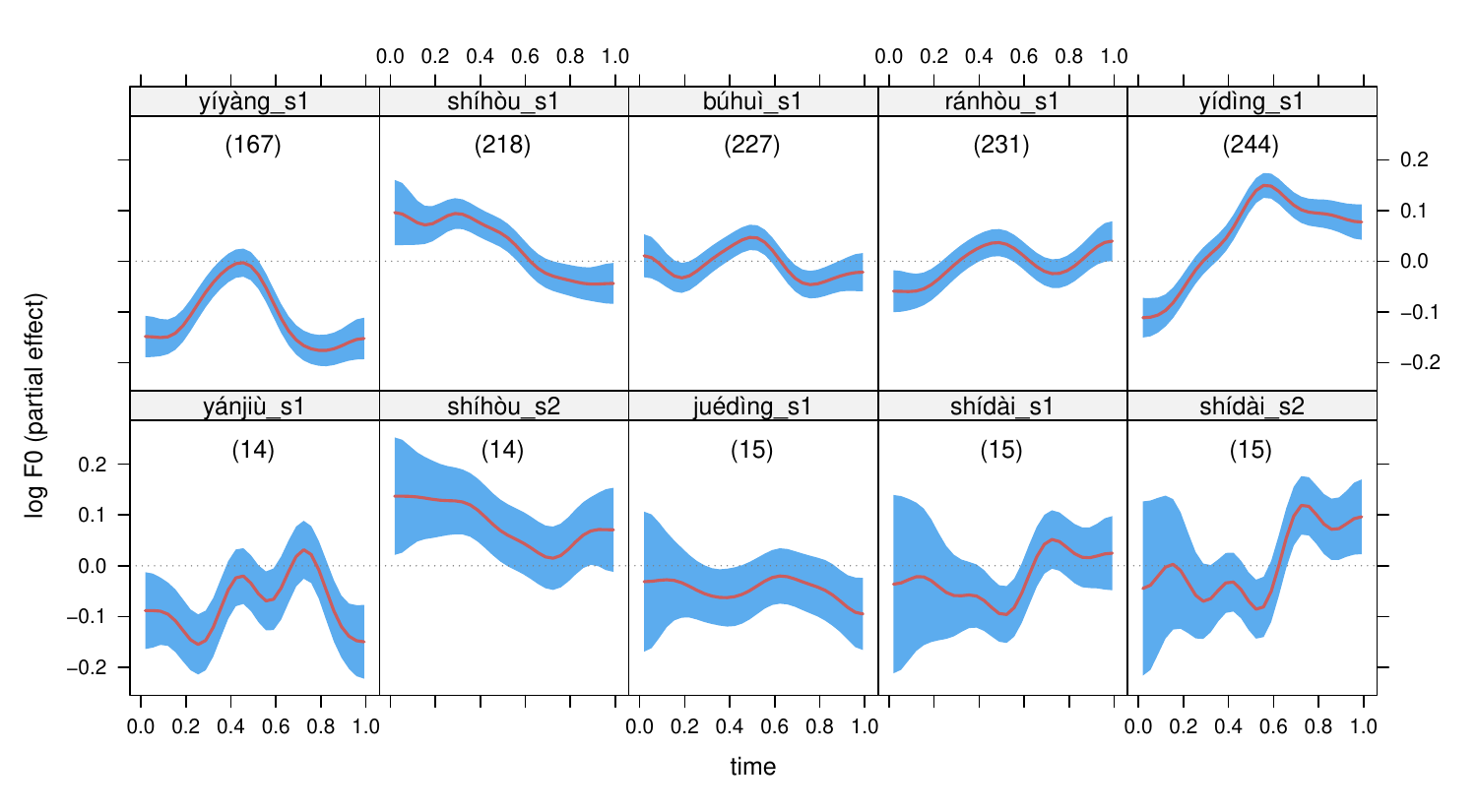}
  \caption{Predicted pitch contours of the partial effects of the factor smooth for \texttt{sense}, for the five most frequent senses (upper row) and the five least frequent senses (lower row). Numbers in parentheses indicate the number of tokens in the dataset for the different senses. 
  }
  \label{fig:sense_err_check}
\end{figure}

There are two reasons for the absence of greater prediction precision for models having access to \texttt{sense} instead of \texttt{word}. Firstly, in the smaller dataset used for these models, no fewer than  35 of the 51 word types are represented by only one sense. Any prediction advantage would therefore have to be contributed by just 16 words.  Secondly, for the majority of this subset of 16 words, one sense accounts for most of the tokens. For tokens with these dominant senses, prediction is possible with greater precision. However, for tokens with less frequent senses, prediction is necessarily less precise. To see this, consider Figure~\ref{fig:sense_err_check}, which presents predicted pitch contours and approximate confidence intervals for senses with many tokens (upper panels) and senses with few tokens (lower panels). Confidence bands are narrower for senses with many tokens. As a consequence, prediction for held-out tokens cannot be of the same quality for senses with few tokens as compared to senses with many tokens. The overall improvement in model fit for the sense-based GAM results from the fact that the pitch contours of tokens with the dominant sense of each word can be better predicted once these tokens are separated from tokens of minority senses.

\subsection{Summary of Section \ref{sec:pitch}} \label{sec:discuss_gam}
The results presented in this section have provided evidence in support of our first two predictions. Firstly, word type is a stronger predictor of tonal realization than all the previously established word-form related predictors combined.
Secondly, information about a word's meaning in context improves prediction of its tonal realization in that context. To the extent that sense labels provide more fine-grained meaning distinctions than word labels do, our results suggest that meaning plays a role in shaping the realization of tonal contours in Mandarin. In other words, in addition to the relevant segmental differences previously identified in the literature, differences in meaning also contribute. 

Nevertheless, as discussed in Section \ref{sec:predictors}, sense labels impose discrete categories on semantic variation that is actually much richer, more subtle and more nebulous than can be captured by such inventories. In the computational models reported in Section \ref{sec:model} below, this problem did not arise since our use of the DLM enabled us to replace relatively crude sense categories with token-specific semantic representations.

\section{Understanding and producing item-specific F0 contours}\label{sec:model}

So far we have shown that it is possible to identify meaning-specific modulations of the pitch contour for Mandarin words with the RF tonal pattern. The question therefore arises as to whether native speakers of Mandarin could in principle profit from these meaning-specific modulations. In other words, are the semantic components in words' pitch contours sufficiently informative that they could facilitate word comprehension for the listener? A related question is whether these subtle semantic modulations are learnable for a speaker, as opposed to arising mechanically each time a word is produced, as one might expect for purely articulatory effects. As outlined in Section \ref{sec:introduction}, our third and fourth predictions, repeated here for convenience, anticipate an affirmative answer to both these questions:

\begin{enumerate}
\setcounter{enumi}{2}
    \item Given a pitch contour, the meaning of its carrier token can be predicted above chance level by a simple computational model with previous experience of that word type.
    \item Assuming it has previous experience of the relevant word type, a simple computational model can produce an appropriate pitch contour for a given meaning.

\end{enumerate}
In this section of the paper, we explore these predictions with computational modeling using the Discriminative Lexicon Model \citep[DLM,][]{baayen2019discriminative,Chuang:Baayen:2021,Heitmeier:Chuang:Baayen:2024}. If we can show that a simple computational model can learn to predict the meaning of a word token from its pitch contour, and that pitch contours can be predicted from intended meaning, we have a proof of concept for the potential functionality of meaning-specific pitch realization in human lexical processing. 

As described in Section \ref{sec:introduction}, the DLM focuses on the relationship between words' forms and their meanings, and allows for fine-grained alignments between low-level features of form and low-level features of meaning. Form-meaning relationships are captured by two networks: a comprehension network that maps word form onto word meaning, and a production network that maps word meaning onto word form.  Recall that in the DLM theory of the mental lexicon, forms and meanings do not have representations in memory. Form representations represent ephemeral auditory or visual input, which dynamically generates a corresponding, equally ephemeral, meaning representation. Conversely, a meaning conceptualized by a speaker at a given point in time is dynamically transformed into ephemeral representations driving articulation. In line with this theory, the DLM generates forms and meanings on the fly on a token by token basis, making it possible to model the relationship between a given token's specific pitch contour and that token's context-specific meaning, and hence to account for correspondences between meaning and fine phonetic detail.\footnote{This is not possible in models of speech production and comprehension that rely on stored abstract representations to mediate between form and meaning, where every token of a given word type is assumed to be associated with the same stored representations \citep[e.g.,][]{Cutler:Clifton:99,levelt1999theory}.}

In the DLM, both the form and the meaning of each word token are operationalized mathematically as high-dimensional numeric vectors. In order to test our predictions about the potential functionality of tonal modulations, the form vectors used in this study are based on the F0 contour of the  relevant token. 
The meaning vectors that we use are context-specific, and hence also vary from token to token. Sections \ref{sec:CE} and \ref{sec:pitchvectors} describe how we obtained the vectors for meaning and pitch, respectively; Section \ref{sec:comprehension} addresses the functionality of pitch in comprehension and \ref{sec:production} does the same for production.

\subsection{Representing meaning: contextualized embeddings}\label{sec:CE}

Embeddings are widely-used numeric representations of words' meanings, developed from the distributional semantic insight that words with similar meanings tend to occur in similar contexts \citep{Firth:1968,Harris1954,Salton1975}. Embeddings represent word meanings as real-valued high-dimensional vectors in a semantic space \citep{Schuetze1992}. They have been found to provide a plethora of novel insights in both psychology \citep{Landauer:Dumais:1997,gunther2019vector,bruni2014multimodal} and linguistics \citep{gahl2022time,perek2017distributional,Nieder:Chuang:vdVijver:Baayen:2023}, and are widely used in computer science \citep{Bojanowski2017}. First generation word embeddings are static, type-level representations that model the meaning of a word type as a fixed point in semantic space, regardless of its usage in a given context. These representations therefore have difficulty distinguishing between multiple senses of a word \citep{Pilehvar2020}, and although various methods have been proposed to incorporate sense or context information into type-level embeddings \citep[see e.g.,][]{Reisinger2010,Huang2012,Neelakantan2014, Iacobacci2015}, most of these methods involve the use of sense-annotated corpora, which as far as we know are not available off the shelf for Mandarin and, in any case, have the disadvantage of discretizing more complex semantic variability. An alternative is to use contextualized embeddings (\textbf{CE}s). In contrast to static, type-level embeddings, which are based on word co-occurrences irrespective of order, CEs take into account the sequence of words in the immediate context of a target word. CEs therefore encode word meanings at the token level, and different tokens of the same word type will have different but similar context-specific embeddings.

To address the issue of words having context-specific meanings, this study used CEs produced for the tokens of our data by a pre-trained unidirectional language model based on the GPT-2 architecture. The model, developed by CKIP, Academia Sinica, Taiwan\footnote{We used \texttt{ckiplab/gpt2-base-chinese}, which is available on \url{https://github.com/ckiplab/ckip-transformers}.}, was trained on a 4.3 billion character dataset written with traditional Chinese characters. The model has 102 million parameters and encodes each character as a 768-dimensional vector. We presented the target words and their preceding contexts (consisting of all the words that occur before the target in the current utterance, as well as all the words in the immediately preceding utterance) to the GPT-2 model, and obtained two embeddings from the model, one for each character. Following standard machine-learning practice \citep[e.g.,][]{huang2021whiteningbert} we then averaged the two embeddings, so that every token in our dataset received a 768-dimensional vector representing its context-specific meaning.

\begin{figure}
  \centering  
  \includegraphics[width=0.8\textwidth]{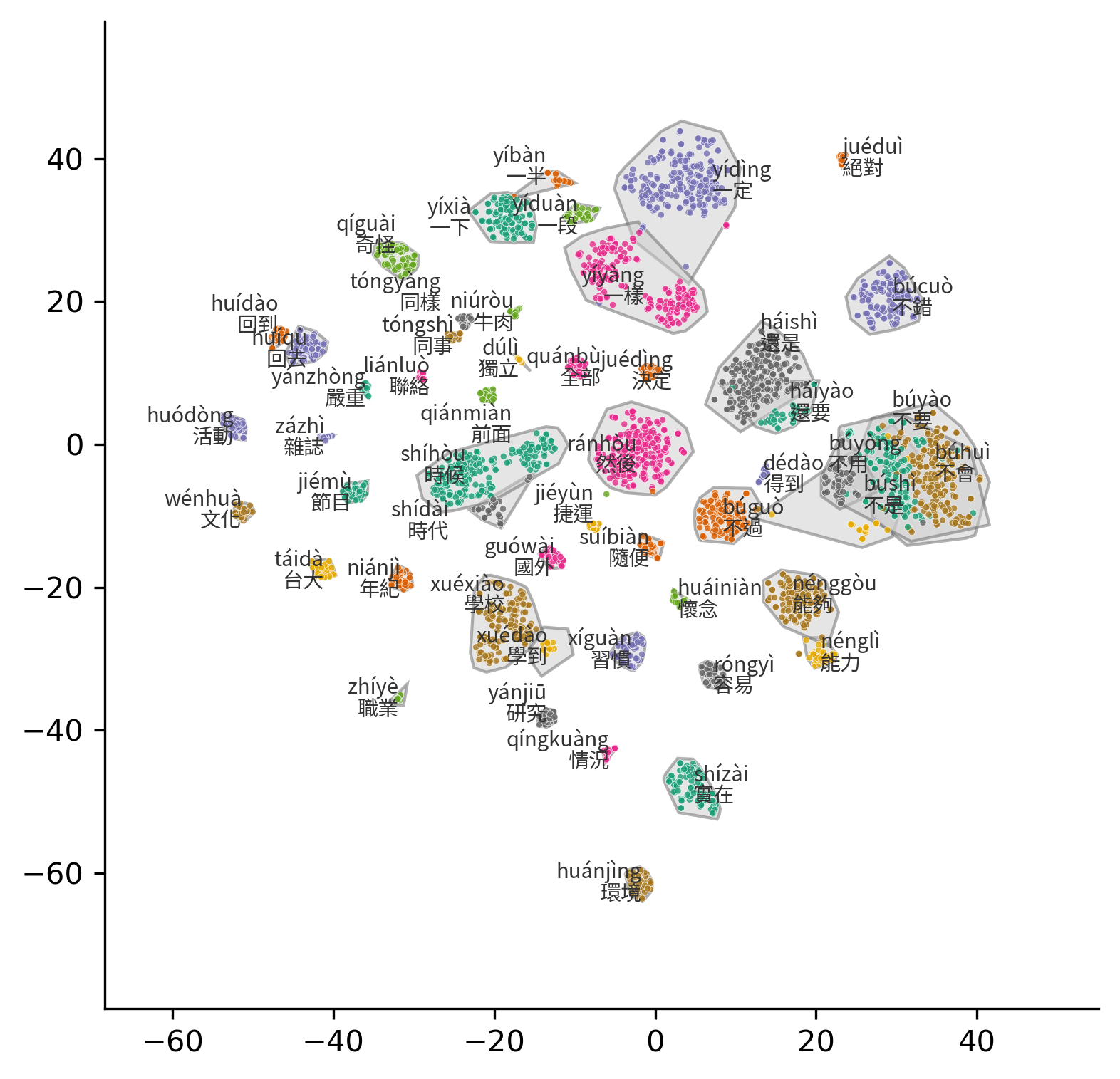}
  \caption{Contextualized embeddings, obtained from a pre-trained Chinese GPT-2 model, cluster by word type in the two-dimensional plane obtained with t-distributed stochastic neighbor embedding \citep{maaten2008visualizing}. Convex hulls (grey polygons) show that the tokens of the different word types form well-localized and highly distinct clusters.}
  \label{fig:CE_token_tsne}
\end{figure}

To visualize the semantic space of the CEs, we reduced the 768-dimensional semantic space to two dimensions using tSNE \citep{maaten2008visualizing}. Figure \ref{fig:CE_token_tsne} shows the resulting reduced 2D plane, using convex hulls to highlight that tokens of different word types typically fall in distinct regions while tokens of the same word type form clear clusters. Although perhaps unsurprising, this distribution confirms that, despite polysemy, the word types in our data do capture a general approximation of what each token means. It is also reassuring to see that, like static embeddings, the CEs can capture inter-word semantic relations. For instance, there is a cluster of school-related words in the lower left: \mandarin{學校} \textit{xue2xiao4} `school', \mandarin{研究} \textit{yan2jiu4} `research', \mandarin{學到} \textit{xue2dao4} `learn+resultative' \citep[see][for similar results]{Vulic2020}.

\subsection{Representing form: pitch vectors}\label{sec:pitchvectors}
\noindent
For the DLM to implement mappings between form and meaning, every form vector input to the model has to have the same number of dimensions as all the others. However, because the tokens in our data vary in duration, our tokens also vary in the number of measurement points. This means that the raw measurements cannot be used to create the form vectors. The raw pitch contours also have the problem that there are gaps due to voicelessness. To overcome these problems, we used two of the GAMs described in Section \ref{sec:pitch} to obtain smoothed pitch contours from which we could extract a standard number of measurements. Although the sense GAM (Section \ref{sec:sense_id}) had a better fit to the data than the word GAM did (Section \ref{sec:word_id}), the former unavoidably used a smaller dataset than the latter. For our DLM models, we wanted to maximize the number of data points available; we therefore chose to use the word model and the corresponding omnibus-segment model. We generated two predicted pitch contours for each token, one using predictions from the word GAM, and the other using predictions from the omnibus-segment GAM. Each of these predicted contours was then used to generate F0 predictions at 50 equally spaced time points ranging between 0 and 1 for every individual token.

\begin{figure}
  \centering
  \includegraphics[width=.8\textwidth]{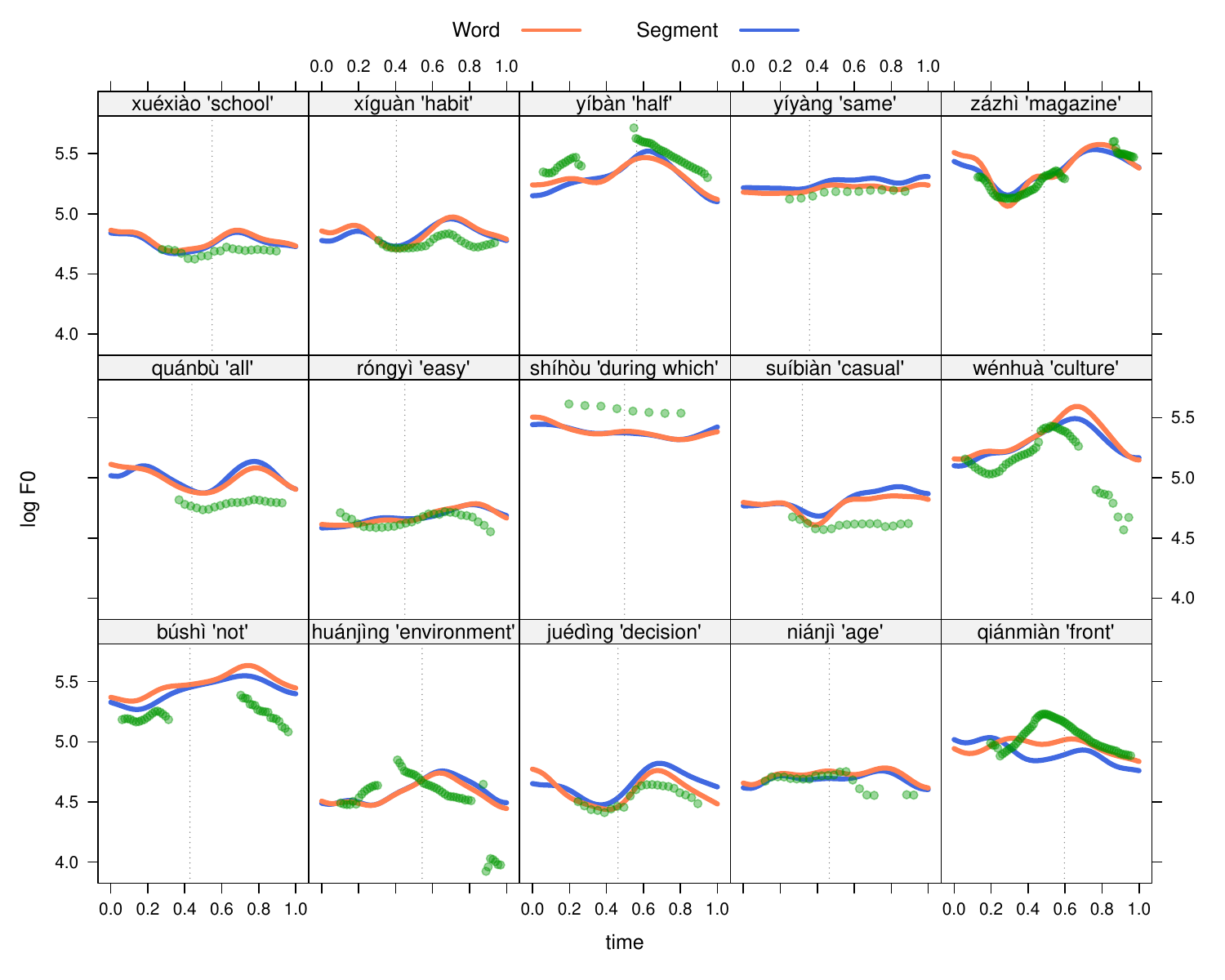}
  \caption{One token randomly selected for a selection of words. The green dots plot the observed pitch contour (raw data), and pitch vectors obtained from the word-type and the omnibus-segment models are represented by the orange and blue curves respectively. The vertical dotted lines indicate syllable boundaries.}
  \label{fig:pitch_vec}
\end{figure}

Both the word GAM and the omnibus-segment GAM include all the speaker-related and context-related control variables described in Section \ref{sec:predictors}. The only difference is that the former additionally includes \texttt{word} as the sole lexical predictor, while the latter includes six predictors specifying words' segmental properties. Examples of GAM-generated contours (from both the word and the omnibus-segment GAMs), together with the raw F0 values, are presented in Figure \ref{fig:pitch_vec}. As can be seen, the GAM-generated contours, though generally smoothing out the undulations in raw F0s, still largely capture the overall contour shape. Moreover, since the two GAMs provide similar but not identical predicted contours, it is possible to compare their performance in the DLM. If pitch contours generated from the word GAM provide superior fits to the respective semantic vectors compared to those generated from the omnibus-segment GAM, this will provide further evidence that the \texttt{word} variable is indeed contributing some meaning-related information.

A speaker's gender and individual characteristics such as vocal tract anatomy, idiolect, and emotional state at the time of speaking, all have strong effects both on their baseline pitch and on pitch range. Similarly, in both our word GAM and in the corresponding omnibus-segment GAM (Section \ref{sec:word_id}), the intercepts are largely dependent on the speaker's gender and individual identity, and differences in amplitude are largely dependent on token duration, which we take to reflect both the speakers' idiolect and their emotional state at the time of speaking, amongst other things. On the semantic side, in normal spoken interaction between humans, a speaker's identity and emotional state not only contribute to the pitch contours they produce, but are also conceptually available to their interlocutors. In contrast, the CEs used as semantic representations in our DLM modeling (Section \ref{sec:CE}) are based entirely on written text and therefore encode much less information about the speaker. To control for this discrepancy, we centered and scaled the predicted F0 values by token; that is, for each token in our data, and for each GAM, we calculated the mean and range of the 50 predicted F0 values, subtracted the mean from each predicted value, and divided the result by the range. This method of scaling (min-max normalization) ensures that the scaled data stays within a fixed range, between 0 and 1, so that every token contributes equally to the model fit, irrespective of its baseline pitch or amplitude; without scaling, tokens with a greater amplitude would be taken into account more than those with a lower amplitude. A consequence of the way we centered and scaled the pitch vectors is that our DLM production models generate predictions for the geometric shapes of the contours, but not for absolute pitch or amplitude.\footnote{Geometric shape can be defined as `all the geometrical information that remains when location, scale, and rotational effects are filtered out from an object' \citep{kendall1977diffusion}.}

\subsection{Modeling comprehension}\label{sec:comprehension}
\subsubsection{Method} 

\noindent
We used two different methods to map our pitch vectors onto our semantic vectors in a comprehension network. The first method involves a straightforward linear mapping using the Linear Discriminative Learning (\textbf{LDL}) engine of the DLM. This is equivalent to the standard linear mappings used in statistics for multivariate multiple regression \citep[see, e.g.,][for introductions]{gahl2022time,heitmeier2021modeling,Heitmeier:Chuang:Baayen:2024}. The second method (henceforth \textbf{ResLDL}) complements the linear mapping with an additional deep mapping, making it possible to accommodate nonlinear relations while keeping the model relatively interpretable. ResLDL augments an LDL mapping with a nonlinear deep network, which is given the task of capturing any systematicities that are left unexplained in the residuals of the linear network (hence the name ResLDL). Using both these methods, and comparing the results, allowed us to shed light on the complexity of the relationship between our pitch vectors and our semantic vectors.\footnote{Technical details of the  LDL and ResLDL implementation can be found in the online supplementary material.}

We split our data into a training set (80\%), a validation set (10\%), and a test set (10\%) in such a way that every word type was represented in all three sets of data and the number of tokens per word was proportional in all three sets. In other words, the test sets consisted entirely of novel tokens but no novel types, simulating the situation for human language use with previous experience. Both the LDL and the ResLDL mappings were trained on the training data and tested on the test data. In accordance with standard machine learning practice, the validation set was used to fine-tune the hyperparameters in the ResLDL model, before testing. This was not necessary for the LDL model since there are no hyperparameters in LDL. To ensure that our results were not specific to a particular data split, we repeated the entire modeling procedure 30 times using repeated training/test splits, i.e., Monte Carlo cross-validation \citep{zhang1993model,kuhn2013applied}. The repeated splits followed the same proportions described above, generating thirty accuracy scores for each combination of pitch type (omnibus-segment or word F0 smooths) and network (LDL or ResLDL).

We evaluated the accuracy of model predictions as follows. For each pitch vector in the test set, we obtained a corresponding predicted semantic vector and identified its closest neighbor among the actual CEs of the tokens in our data. If this nearest neighbor belonged to any token of the same word type as the target token, the predicted semantic vector was assessed as correct, and otherwise as incorrect. This measure of success was chosen for both computational and conceptual reasons, as detailed in the following two paragraphs. 

Although one might expect that a predicted semantic vector would ideally be closest in semantic space to the CE of the held-out token in question, this is computationally unrealistic. The CEs in our models are conditioned on the preceding context of a given token, and are uninformed about the following context. The pitch contours, in contrast, are shaped in part by the tone on the following word, and the probability of the word given the next word. Thus, there is information in the pitch contours that is absent in the CEs, making it computationally infeasible to predict token-specific vectors. Furthermore, both the pitch contours and the CEs have measurement error. Similar to the way that a linear regression line predicts the mean value of a dependent variable for a given value of an independent variable, but not the individual data points used to generate the line, here we can predict at the level of types, but not at the level of individual tokens.

From a cognitive perspective, it is worth noting that listeners cannot arrive at exactly the same conceptualisation as the speaker, as listeners and speakers have different experiences with the language and different life histories. For example, a listener who hears `Do you fancy a coffee?' may conceptualize a cappuccino, even if the speaker was envisioning an espresso. Fortunately, provided both interlocutors arrive at similar enough meanings, communication can proceed unhindered. In addition to being computationally feasible, using same word type as a criterion for success therefore makes sense in terms of human performance levels.

\subsubsection{Results} 
\begin{figure}
\centering
\includegraphics[width=0.7\textwidth]{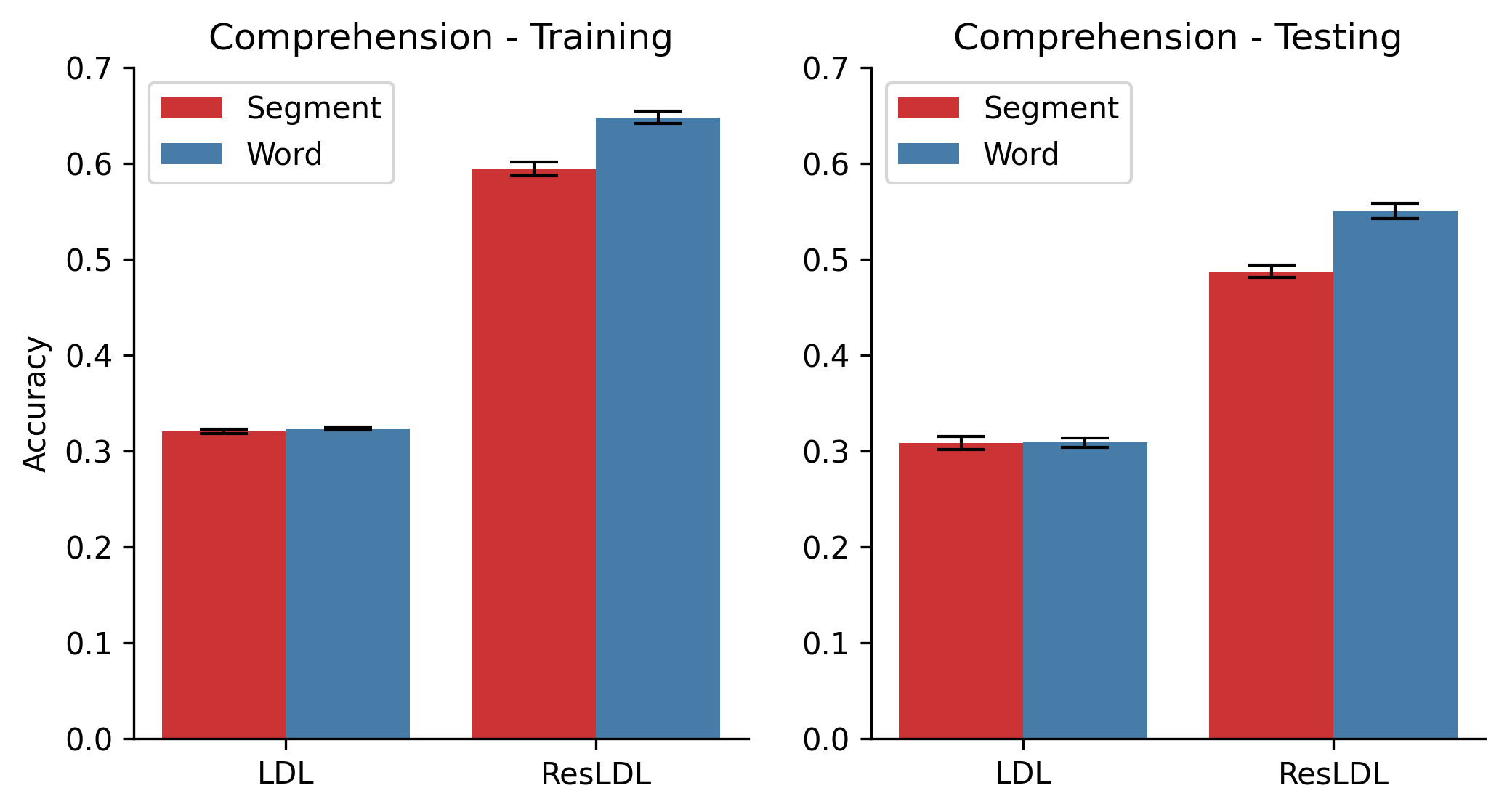}
\caption{Mean comprehension accuracies for training data (left panel) and test data (right panel) for LDL and ResLDL mappings from omnibus-segment (red) and word (blue) pitch vectors. Mean accuracy is obtained from 30 stratified random training and testing splits, each trained and evaluated independently. Error bars indicate double the standard error.}
\label{fig:top-word-comp}
\end{figure}

Figure~\ref{fig:top-word-comp} presents the mean comprehension accuracies for the training data (left) and the test data (right). The individual barplots show accuracies for LDL (left two bars) and ResLDL (right two bars), for pitch contours based on segment-aware GAMs (red) and pitch contours based on word-aware GAMs (blue). As mentioned above, a given prediction is considered correct when the closest neighbor of the predicted CE is of the same word type as the target. For LDL, accuracy hovers around 30\% for both training and test data, whereas for ResLDL accuracy is higher, over 60\% and 50\% for training and test data respectively. These results are surprisingly good, given that the models are requested to predict semantic vectors on the basis of pitch information only, notwithstanding the fact that pitch contains implicit information about phonological segments (cf. Section \ref{sec:eval_word}). For comparison, across our whole dataset the theoretical probability of a pitch vector and CE belonging to the same word type by chance is approximately 0.038. Similarly, baseline accuracies obtained by evaluating on a dataset with randomly permuted word labels were 3.7\% for the training set, and 3.5\% for the test set. This allows us to conclude that the classification accuracies of our models are far from trivial. On the contrary, even the least successful model achieves accuracies that are a whole order of magnitude greater than would be expected by chance. 

The higher accuracies of the ResLDL model compared to the LDL model indicate that mappings from pitch contours to CEs have significant nonlinear components. This nonlinear mapping may be required because we are mapping from 50-dimensional pitch contours to 768-dimensional CEs. As it is impossible to map a lower-dimensional space into a higher-dimensional space with a linear mapping without losing information,\footnote{
To see this, consider points in a cube, and their projection onto a plane in that cube. From that projection, which is two-dimensional, the original locations of the points in the three-dimensional cube cannot be fully reconstructed.} the greater accuracy of ResLDL is unsurprising. Nevertheless, it is remarkable that the linear mappings show very similar performance on training and test data, suggesting that there is a strong linear component to predicting meaning from tonal contours.

A comparison of results from the two types of pitch vectors shows that meaning prediction with ResLDL is more accurate when the pitch contour smooths are generated using the word GAM than using the omnibus-segment GAM. This indicates that the factor smooth for \texttt{word} not only contributes to a better model fit in the GAM, but also produces predicted pitch contours that are better aligned with words' meanings, albeit in a nonlinear way. Of course, \texttt{word} encodes all the information about words' segmental makeup that is given to the omnibus segment model, so the resulting contours do contain this information. But the superior performance of the word-based contours shows that tonal realizations that include \emph{all} information associated with word type have the potential to help listeners to identify words' meanings even more accurately than contours with just segmental information.

\subsection{Modeling production}\label{sec:production}

\subsubsection{Method} 
We have seen that the pitch contours of Mandarin disyllabic words contain substantial information about word meaning. It is remarkable that a DLM comprehension model can achieve a test accuracy of over 50\% when modeling with word-aware pitch contours and ResLDL. We now turn to production, addressing the question of whether a token's pitch contour can be predicted with reasonable accuracy from its CE. If so, this would support our hypothesis that speakers can in principle learn to produce meaning-specific tonal contours.

Before going into further detail, we note that this task is considerably more difficult than the task presented to the word GAM model in Section \ref{sec:pitch}. The GAM model was asked to predict pitch contours from word labels and was oblivious to variation in meaning between tokens of a given word type. In the models reported below, however, the LDL and ResLDL mappings are confronted with semantic vectors that are different from token to token. The question is whether the similarities between the CEs of tokens belonging to the same word are sufficiently consistent for the LDL and ResLDL mappings to predict appropriately similar pitch contours.

Model set-up was the same as for comprehension, except that to model production the input consisted of CEs and the output consisted of pitch vectors. We again conducted the modeling procedure 30 times. 
For each CE in the test set, we obtained a corresponding predicted pitch vector and identified its closest neighbor among the actual (GAM-generated) contours of the tokens in our data. If this nearest neighbor belonged to any token of the same word type as the target token, the predicted pitch vector was assessed as correct, and otherwise as incorrect.

We complemented the quantitative evaluation with a qualitative analysis of the pitch contours predicted by the model for individual word types. To do this, we calculated the centroid of the CEs for all tokens of a given word, and used this centroid vector to generate a predicted pitch contour from the production network with LDL mappings and word-based pitch vectors. For each of the words presented in Figure~\ref{fig:word_pred}, we then assessed the quality of this LDL-predicted contour by visually comparing it with the contour produced by averaging the actual (GAM-generated) pitch vectors used to train the model, for all tokens of the word in question.

\subsubsection{Results}

\begin{figure}
\centering
\includegraphics[width=0.7\textwidth]{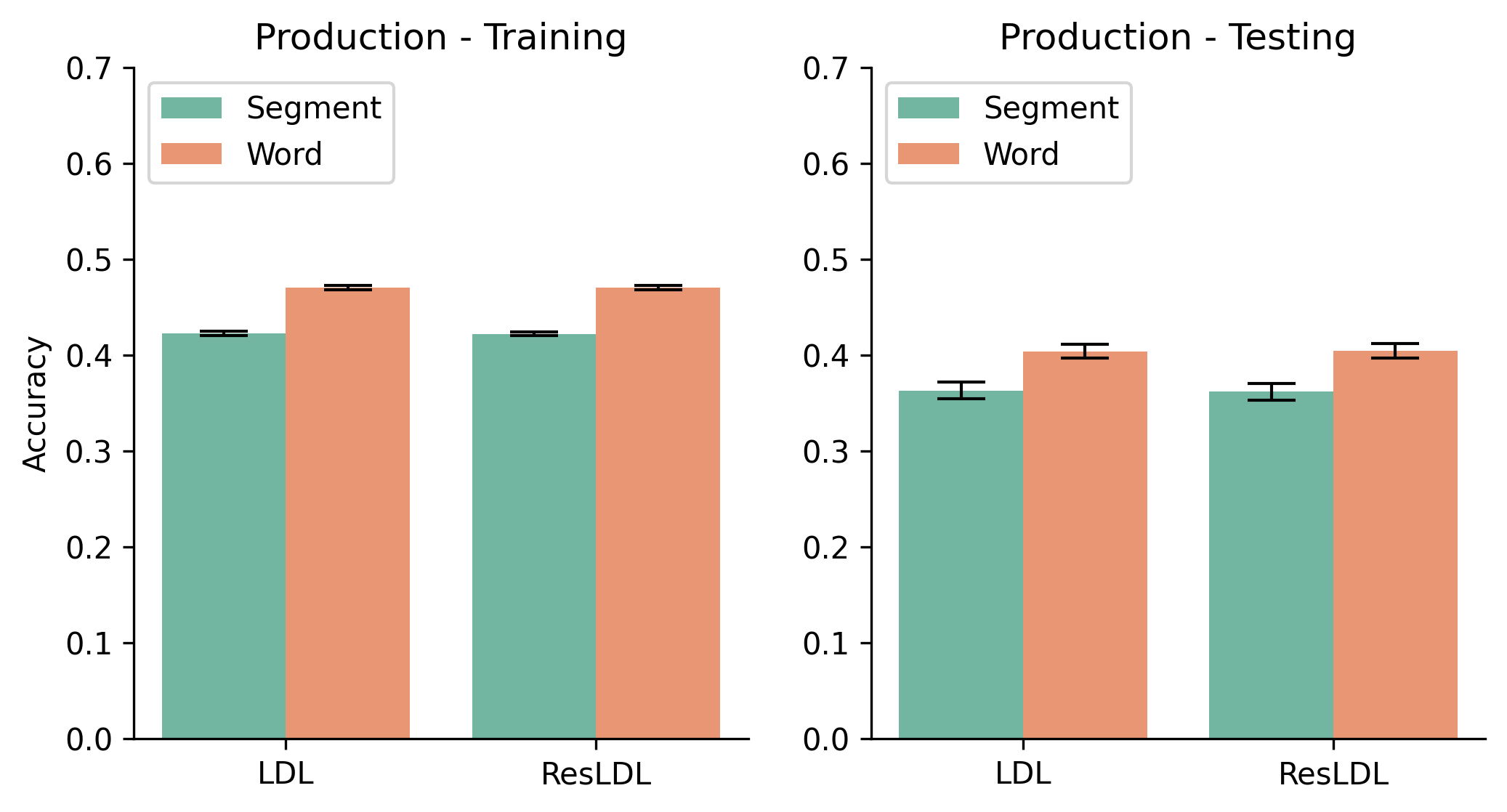}
\caption{Mean production accuracies for training data (left panel) and test data (right panel) for LDL and ResLDL mappings from omnibus-segment (green) and word-type (orange) pitch vectors. Mean accuracy is obtained from 30 stratified random training and testing splits, each trained and evaluated independently. Error bars indicate double the standard error.}
\label{fig:top-word-prod}
\end{figure}

\noindent
Mean production accuracies (over 30 repetitions) for the token-based evaluation are presented in Figure~\ref{fig:top-word-prod}. For training data (left), accuracies are between 40\% and 50\%. The accuracies for the test data are only slightly lower, hovering around 40\%.  The probability of a CE and pitch vector belonging to the same word type by chance is the same as for the comprehension models, namely 0.038. Permutation baselines are again 3.7\% for training and 3.5\% for testing. In other words, like the comprehension models, the production models have accuracies an order of magnitude greater than would be expected due to chance. However, in contrast to the comprehension results, production accuracies are remarkably similar for LDL and ResLDL. Apparently, linear mappings suffice when predicting low-dimensional pitch contours from high-dimensional CEs, and succeed in capturing the regularities in the meaning-to-form mappings. Possibly, predicting pitch from semantics is a cognitively more natural task than predicting semantics from just pitch on its own, and hence requires less powerful mappings. Finally, as for comprehension mappings, predicting pitch contours from CEs is more successful when pitch contours are generated with word-based GAMs, compared to segment-based GAMs.

\begin{figure}
  \centering
  \includegraphics[width=0.8\textwidth]{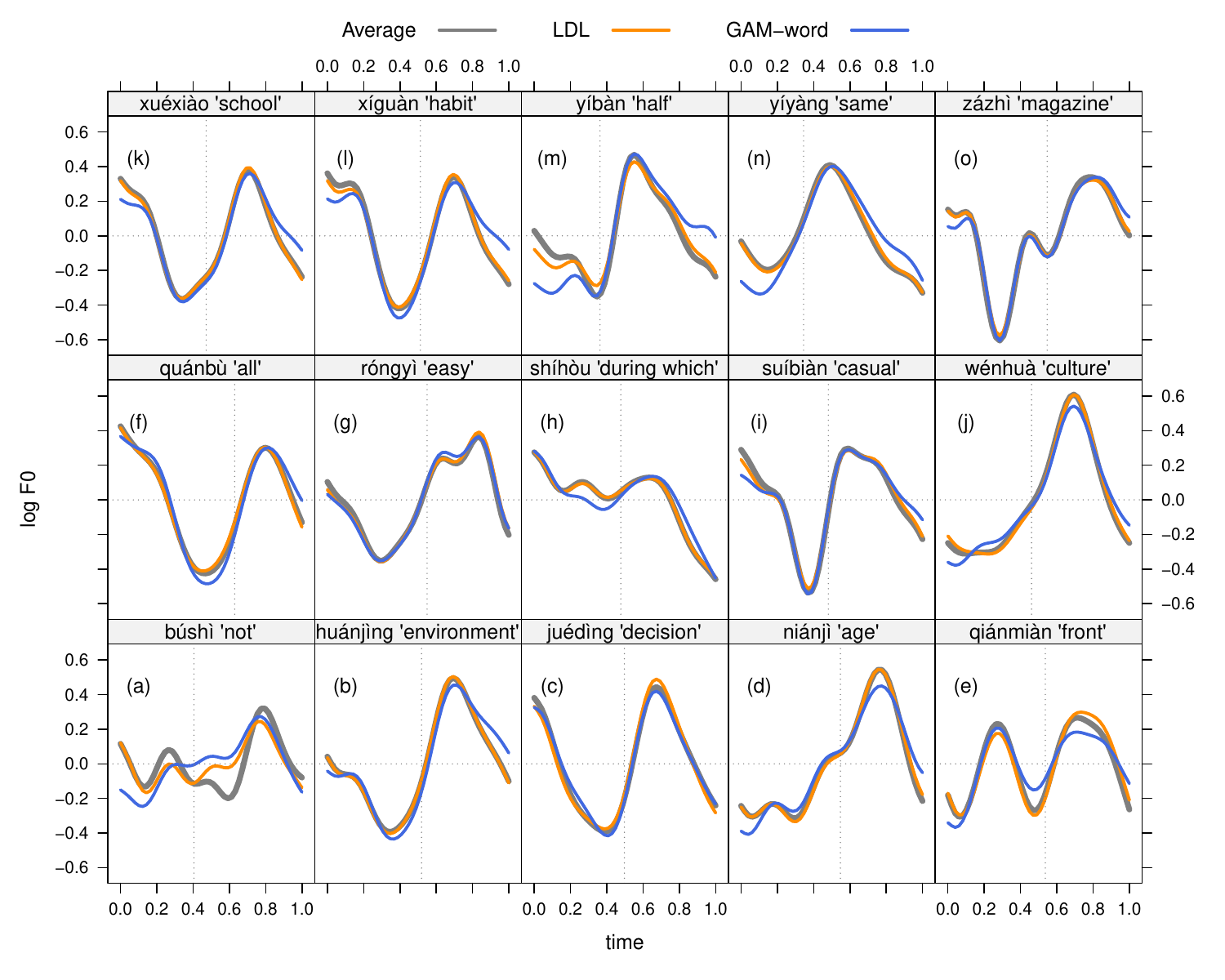}
  \caption{Pitch contours for the sample of 15 word types introduced in Figure \ref{fig:word_pred}. The gray lines represent the average of the pitch vectors generated by the word-type GAM across all tokens of that type, i.e. the average of the contours used to train LDL. The orange lines represent the predictions generated by LDL. These LDL contours were predicted from `centroid' word meaning, obtained by averaging the CEs of all tokens of the same type. The blue lines represent the word-specific contours predicted by the word GAM as presented in Figure \ref{fig:word_pred} and reproduced here after centering and scaling, i.e., these blue lines show the pure effect of word on the pitch contour, irrespective of other predictors. The vertical dotted lines in the panels indicate the average (word-specific) syllable boundary.}
  \label{fig:word_mod}
\end{figure}

The results of the qualitative analysis are shown in Figure~\ref{fig:word_mod}. The LDL-predicted contours are shown in orange, and the by-type averages of the contours used to train the model are shown in gray. A comparison of these two contours for any given word reveals remarkable similarity, indicating that the LDL production model generates high quality predictions for the shapes of the pitch contours. It is also striking that, in shape, these orange and grey contours closely resemble the contours in Figure~\ref{fig:word_pred}, reproduced for convenience as the blue contours in Figure~\ref{fig:word_mod}.\footnote{Note that the blue contours in Figure~\ref{fig:word_mod} are identical to the contours in Figure~\ref{fig:word_pred}, except that the amplitudes are different due to the way we centered and scaled the contours in Figure~\ref{fig:word_mod}.} Recall from Section \ref{sec:eval_word} that this third set of contours was produced by combining the partial effect smooth for each word type with the general smooth for time for female speakers. The similarity therefore suggests that the word-specific pitch contours isolated by our word GAM can be understood as pitch contours that correspond to the centroids of word's contextualized embeddings.
From this, we draw the conclusion that there is considerable isomorphy between the space of token-specific pitch contours and the semantic space of token-specific embeddings.

In addition to by-word centroids, we also calculated the centroid of the CEs for all tokens of all types in our dataset. The pitch contour predicted from this overall centroid is very similar to the pitch contour in the right-hand panel of Figure~\ref{fig:toy_contour}. 
We therefore infer that the centroid of the embeddings of all tokens can be interpreted as the `meaning' of the unmodulated rise-fall pitch contour. The 10 tokens that are closest to this centroid belong to the word types \mandarin{不過} \textit{bu2guo4} `but, however', \mandarin{然後} \textit{ran2hou4} `and then', and \mandarin{時候} \textit{shi2hou4} `during which', which suggests that these words are the most typical carriers of the RF pitch contour in the current dataset. 
 
\section{General discussion}\label{sec:discussion}

This study investigated variation in the F0 contours of disyllabic words with the rising-falling (RF) tonal pattern in Taiwan Mandarin. The central hypothesis of our study is that Taiwan Mandarin disyllabic word tokens have pitch contours that are in part driven by their meanings. In standard analyses of tone in Mandarin, the rising tone of the first syllable and the falling tone of the second syllable of RF words are inherited from the single-syllable constituents and are taken to be basic, underlying tones. Deviations from these tones are explained by appealing to articulatory and prosodic constraints governing how tones can be realized. Our hypothesis adds word meaning as a missing player in the articulatory arena by arguing that meaning co-determines the realization of Mandarin tones.

Our core hypothesis generates four predictions. The first prediction is that word type will be a stronger predictor of tonal realization than all the previously established word-form related predictors combined. This prediction follows from the hypothesis because word type includes information about meaning in addition to information about form. 
Using generalized additive models (GAMs), we were able to show that word type is indeed a more powerful predictor of tonal realization than a wide range of words' form properties considered jointly. We not only established that the GAM with a factor smooth for word type provided a substantially improved model fit, but also demonstrated that the word-informed GAM provided more accurate prediction for the F0 contours of held-out data. We concluded that individual word types have specific properties --- over and above their segmental makeup --- that modulate the general, sine-wave shaped F0 contour characteristic of words described as having a rising-falling tone pattern.  These specific properties, we conjectured, are semantic in nature.

The second prediction is that information about a word's meaning in context will improve prediction of its tonal realization. If words with the very same segments and canonical tones, but different meanings, have distinct tonal realizations, this provides evidence for the possibility that there is a semantic component to the realization of tone in Taiwan Mandarin disyllabic words. Again using generalized additive modeling, we were able to show that adding information about meaning in context, i.e., sense, did lead to significant improvement in model fit. 
These results provide evidence for the possibility that Mandarin disyllabic word tokens indeed have tonal realizations that are partially determined by their semantics.

The third prediction is that given a pitch contour, the meaning of its carrier token can be predicted above chance level by a simple computational model with previous experience of that word type. To test this, we used the framework of the Discriminative Lexicon Model.  Given the difficult task of predicting words' high-dimensional semantic embeddings from low-dimensional pitch contours,  the DLM comprehension model that we implemented performed on held-out data with an accuracy of over 50\%, compared with a random baseline of 3.5\%. The tonal contours of word tokens turned out to be far more revealing about their meanings than anything we thought might be possible when we started this investigation.

This finding has two important implications. Firstly, for the model to learn to predict meanings from pitch contours, pitch contours must contain information that aligns with aspects of word meaning. This suggests that human speakers of Mandarin could also potentially make use of  this information to optimize speech comprehension.
Secondly, meaning-specific pitch contours might be related to the extensive homophony in Mandarin. According to the Chinese Lexical Database \citep{sun2018chinese}, about 90\% of monosyllabic Mandarin words have at least one homophonous counterpart. From a functional perspective, the presence of meaning-specific pitch modulations may therefore compensate for the lack of semantic discriminability afforded by segmental makeup and syllable structure \citep[see, e.g.,][for a discussion of theoretical implications]{sampson2015chinese,sampson2019unaddressed}. However, given that the present study focuses on disyllabic words, which exhibit much less homophony than monosyllabic words, it remains an empirical question whether word-specific pitch contours for disyllabic words have the same functional load as contours for monosyllabic words would have. We leave this interesting question to future studies.

The fourth and final prediction that follows from our central hypothesis is that, assuming it has previous experience of the relevant word type, a simple computational model can produce an appropriate pitch contour for a given meaning. We again tested this prediction using the DLM. A network trained to match context-specific semantic embeddings to token-specific pitch contours performed far above a random baseline, with accuracies ranging from 40\% to 45\% on training data, and 35\% to 40\% on testing data. Given that the computational models were forced to predict pitch across tokens produced by many different speakers, without any information about the segmental make-up or syllable structure of the words, this is a remarkable result that provides strong support for human speakers in principle being able to learn to produce meaning-specific pitch contours. At this point, however, a word of caution is appropriate. Our DLM models were given the task of predicting the geometric shape of pitch contours, and did not address token- and word-specific differences in pitch height and amplitude. The modeling of the full pitch contours, including height and amplitude, is left for future investigation.

Although the four predictions that follow from our central hypothesis are empirically well-supported, this does not necessarily imply that our hypothesis is correct. 
We cannot rule out that the importance of meaning in our models might actually be due to factors that we did not take into account in our analyses. For instance, the effects of prosody, pragmatics, syntax, and emotion could, in principle, conspire to yield effects that would seem to imply token-specific semantic effects. Measures of surprisal and informativity other than the forward and backward probabilities that we included as control variables may also be informative \citep{tang2021prosody}. In addition, it could be argued that in the present study, which is based on spontaneous conversational speech, the consequences of contraction and reduction \citep{Ernestus:2000,Johnson:2004,tseng2005contracted} are not controlled for. And indeed, we agree, all these factors are worth further investigation. However, it seems unlikely to us that any of these factors will turn out to explain away completely the effect of meaning. Our analyses are based on multiple tokens of each word type, that vary with respect to their syntactic position, their pragmatic function, the amount of segmental reduction, and their emotional valence. The pitch modulations estimated with the help of the GAM models are statistical generalizations across all this variation that is present in our data. It is unlikely that the factors we were not able to control for will be distributed across our tokens in such an unbalanced way that they would be able to explain away our semantic effects. But even were it to turn out that tonal variation is completely predictable from factors other than semantics, our key point is still valid: our DLM models show that this word-specific variation is beautifully fine-tuned with words' meanings as represented by contextualized embeddings.

In our DLM comprehension model, the mapping from a pitch contour to its context-specific embedding is to a large extent linear; in the production model, the reverse mapping is completely linear. These facts indicate that there is considerable isomorphy between the form space of Mandarin word tokens' pitch contours and the semantic space of Mandarin words' context-specific meanings. In other words, form and meaning mirror each other to a much greater extent than is often assumed, especially in frameworks that take as axiomatic that language has a `dual articulation' \citep{Martinet:65} that allocates form and meaning to two unrelated, orthogonal, components of the grammar. Furthermore, the existence of this isomorphy means that human language users could in principle exploit the associations between form and meaning to optimize comprehension and production.

The question then arises as to whether listeners actually do make use of the distributional-statistical information that is in the speech signal of Mandarin RF words: we think there is a strong possibility that they do so. The pertinent information is present in the speech signal, so speakers are producing tonal realizations that align with word meaning. But this isomorphy cannot be reduced to the consequences of bio-mechanical constraints on the speech production process. Listeners must be learning the distributional statistics of tone and meaning from the speech to which they are exposed. We hasten to note that the learning of the systematicities between pitch and semantics is in all likelihood a completely subliminal process.  It is not necessary for learners to be aware of the subtle modulations of pitch contours in relation to equally subtle nuances in meaning.  In our conception of the learning process, successful understanding, token by token, will drive low-level learning in the lexical networks, without conscious reflection and effort being required.\footnote{For token-by-token incremental lexical learning, see \citet{heitmeier2023trial}, and for continuous recalibration in vision, see \citet{Marsolek:2008}.
}

To conclude, we have provided a range of observations that are consistent with the possibility that the details of how tones are realized in Taiwan Mandarin disyllabic words is partially determined by meaning in context. If our interpretation of these observations is on the right track, semantics is an important missing player in current phonetic studies of F0 modulation in Mandarin. We believe our empirical findings are sufficiently strong to open up new lines of research on the realization of pitch in tone languages. We also hope that our findings will contribute to an improved understanding of why deep learning speech processing systems are so remarkably effective and now constitute the state-of-the-art in natural language processing. Our hypothesis is that these systems can pick up systematicities between form and meaning that are not open to human introspection, but that are visible to GAMs and computational models. Crucially, we hypothesize that these systematicities are not only exploited by computational modeling algorithms, but are also essential, albeit subliminally, for optimizing human lexical processing in comprehension and production.


\clearpage
\bibliography{bib_arXiv}

\end{document}